\documentclass{article} % For LaTeX2e

\PassOptionsToPackage{pdfusetitle}{hyperref}

\usepackage{iclr2026_conference,times}

\usepackage{amsmath,amsfonts,bm}

\def\figref#1{figure~\ref{#1}}
\def\eqref#1{equation~\ref{#1}}
\def\1{\bm{1}}

\def\vtheta{{\bm{\theta}}}

\def\vh{{\bm{h}}}

\def\vt{{\bm{t}}}

\def\vx{{\bm{x}}}

\def\vz{{\bm{z}}}

\def\mQ{{\bm{Q}}}

\DeclareMathAlphabet{\mathsfit}{\encodingdefault}{\sfdefault}{m}{sl}
\SetMathAlphabet{\mathsfit}{bold}{\encodingdefault}{\sfdefault}{bx}{n}

\usepackage{booktabs}                   % professional-quality tables
\usepackage{nicefrac}                   % compact symbols for 1/2, etc.
\usepackage{microtype}                  % microtypography
\usepackage{url}                        % simple URL typesetting
\usepackage{caption}
\usepackage{bookmark}                   % add PDF bookmarks
\usepackage{wrapfig}
\usepackage{float}

\usepackage{amsmath}
\usepackage{amssymb}
\usepackage{amsfonts}
\usepackage{amsthm}
\usepackage{mathtools}

\usepackage[mode=match,uncertainty-mode=separate,detect-weight=true,retain-zero-uncertainty=true]{siunitx}

\usepackage[ruled]{algorithm2e}
\DontPrintSemicolon
\usepackage{algpseudocode}

\usepackage{multirow} 
\usepackage{makecell}

\usepackage[pdfusetitle]{hyperref}
\usepackage{xcolor}
\usepackage{xkcdcolors}
\hypersetup{
  pdfauthor={David Lüdke, Marten Lienen, Marcel Kollovieh, Stephan Günnemann},
  pdfsubject={ICLR 2026},
  colorlinks=true,
  linkcolor=xkcdCrimson,
  urlcolor=xkcdBluish,
  citecolor=xkcdLightNavy,
}

\usepackage{enumitem}
\setlist[itemize]{leftmargin=30pt,rightmargin=25pt}

\makeatletter
\patchcmd\WF@putfigmaybe{\lower\intextsep}{}{}{\fail}%
\AddToHook{env/wraptable/begin}{\setlength{\intextsep}{0pt}}
\makeatother

\usepackage[status=draft,layout=inline,author=TODO]{fixme}
\fxusetheme{color}

\usepackage[capitalise]{cleveref}

\usepackage[
  acronym,
  style = long,
  nopostdot,
]{glossaries}
\glsdisablehyper{}

\newacronym[longplural=temporal point processes]{tpp}{TPP}{temporal point process}
\newacronym[longplural=intensity-free TPPs]{iftpp}{IFTPP}{intensity-free TPP}
\newacronym{ctmc}{CTMC}{continuous-time Markov Chain}
\newacronym{mmd}{MMD}{maximum mean discrepancy}
\newacronym{mre}{MRE}{mean relative error}
\newacronym{mlp}{MLP}{multi-layer perceptron}
\newacronym{ema}{EMA}{exponential moving average}

\newcommand{\ours}{\textsc{EdiTPP}}
\newcommand{\psdiff}{\textsc{PSDiff}}
\newcommand{\addthin}{\textsc{AddThin}}
\newcommand{\iftpp}{\textsc{IFTPP}}

\DeclareMathOperator{\rmblanks}{f_{rm-blanks}}
\DeclareMathOperator*{\E}{\mathbb{E}}
\DeclareMathOperator{\ins}{ins}
\DeclareMathOperator{\sub}{sub}
\DeclareMathOperator{\del}{del}
\DeclareMathOperator{\alignseqs}{align}
\newcommand{\model}{u_{s}^{\vtheta}}
\newcommand{\pnoise}{p_{\text{noise}}}
\newcommand{\pdata}{q_{\text{data}}}
\newcommand{\dataspace}{\mathcal{X}}
\newcommand{\auxspace}{\mathcal{Z}}
\newcommand{\insbins}{b_{\text{ins}}}
\newcommand{\subbins}{b_{\text{sub}}}
\newcommand{\tppsupport}{\mathcal{T}}
\newcommand{\statespace}{\dataspace_{\tppsupport}}
\newcommand{\uniform}{\mathcal{U}}

\newcommand{\operation}{\omega}
\newcommand{\operations}{\Omega}
\newcommand{\setgenerated}{\mathcal{X}}
\newcommand{\setdata}{\mathcal{X}'}
\newcommand{\dxiao}{d_{\mathrm{Xiao}}}
\newcommand{\diet}{d_{\mathrm{IET}}}
\newcommand{\dlen}{d_{l}}

\newcommand\extrafootertext[1]{%
  \bgroup%
  \renewcommand\thefootnote{\fnsymbol{footnote}}%
  \renewcommand\thempfootnote{\fnsymbol{mpfootnote}}%
  \footnotetext[0]{#1}%
  \egroup%
}

\title{Edit-Based Flow Matching for Temporal \\Point Processes}

\author{David L\"udke\textsuperscript{*}, Marten Lienen\textsuperscript{*}, Marcel Kollovieh, Stephan G\"unnemann \\
\texttt{\{d.luedke,m.lienen,m.kollovieh,s.guennemann\}@tum.de} \\
School of Computation, Information and Technology \& Munich Data Science Institute\\
Technical University of Munich
}

\iclrfinalcopy % Uncomment for camera-ready version, but NOT for submission.

\begin{document}

\extrafootertext{\textsuperscript{*}Equal contribution}
\extrafootertext{\phantom{\textsuperscript{*}}Find the code at \href{https://cs.cit.tum.de/daml/editpp}{cs.cit.tum.de/daml/editpp}}

\maketitle

\begin{abstract}
	\looseness=-1
	Temporal point processes (TPPs) are a fundamental tool for modeling event sequences in continuous time, but most existing approaches rely on autoregressive parameterizations that are limited by their sequential sampling.
	Recent non-autoregressive, diffusion-style models mitigate these issues by jointly interpolating between noise and data through event insertions and deletions in a discrete Markov chain.
	In this work, we generalize this perspective and introduce an Edit Flow process for TPPs that transports noise to data via insert, delete, and substitute edit operations.
	By learning the instantaneous edit rates within a continuous-time Markov chain framework, we attain a flexible and efficient model that effectively reduces the total number of necessary edit operations during generation.
	Empirical results demonstrate the generative flexibility of our unconditionally trained model in a wide range of unconditional and conditional generation tasks on benchmark TPPs.

	%Recent work has proposed non-autoregressive diffusion-style generative models for TPPs, to interpolate between data and noise processes in a discrete Markov chain, by deleting noise events and inserting data events over a fixed number of steps.
\end{abstract}

\section{Introduction}\label{sec:introduction}

\begin{wrapfigure}[18]{r}{0.45\textwidth}
	% \begin{minipage}{0.5\textwidth}
	\vspace{-0.2cm}
	\centering
	\includegraphics[width=\linewidth, trim={4.3cm 4.9cm 14.cm 0.8cm},clip]{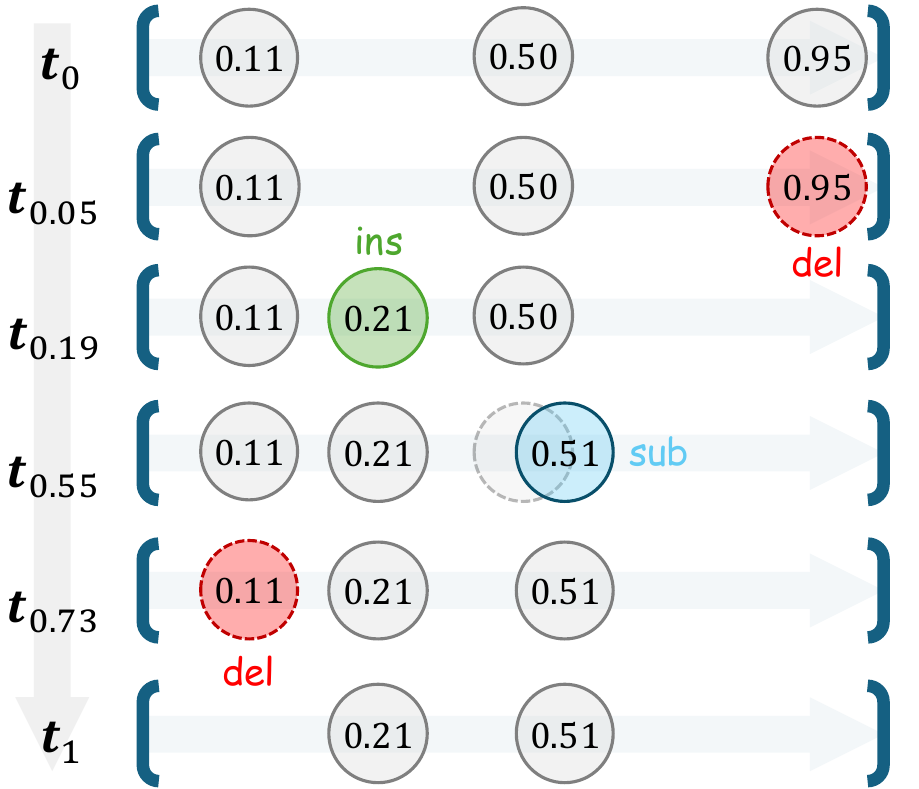}
	\vspace{-0.27cm}
	\caption{Edit process transporting $\vt_0\sim p_{\text{noise}}(\vt)$ to $\vt_1\sim q_{\text{target}}(\vt)$ by inserting, deleting and substituting events.}\label{fig:figure1}
	% \end{minipage}
\end{wrapfigure}

\Glspl{tpp} capture the distribution over sequences of events in time,
where both the continuous arrival-times and number of events are random.
They are widely used in domains such as finance, healthcare, social networks, and transportation, where understanding and forecasting event dynamics and their complex interactions is crucial.
Most (neural) \glspl{tpp} capture the complex interactions between events \emph{autoregressively}, parameterizing a conditional intensity/density of each event given its history \citep{daley2006introduction,shchur2021neural}.
While natural and flexible, this factorization comes with inherent limitations: sampling scales linearly with sequence length, errors can compound in multi-step generation, and conditional generation is restricted to forecasting tasks.

\textbf{Beyond autoregression.}
Recent advances demonstrate that modeling event sequences \emph{jointly} proposes a sound alternative to overcome these limitations.
Inspired by diffusion, \addthin\ \citep{addthin} and \psdiff\ \citep{psdiff} leverage the thinning and superposition properties of \glspl{tpp} to construct a discrete Markov chain that learns to transform noise sequences $\vt_0\sim p_{\text{noise}}(\vt)$ into data sequences $\vt_1\sim q_{\text{target}}(\vt)$ through \emph{insertions} and \emph{deletions} of events.
These methods highlight the promise of joint sequence modeling for \glspl{tpp} by learning stochastic set interpolations and have shown state-of-the-art results, especially in forecasting.

In parallel, \citet{editflow} introduced Edit Flow, a discrete flow-matching framework \citep{gat2024discreteflowmatching,campbell2024generativeflowsdiscretestatespaces,shi2025simplifiedgeneralizedmaskeddiffusion} for variable-length sequences of tokens (e.g., language).
Their approach models discrete flows in sequence space through \emph{insertions}, \emph{deletions}, and \emph{substitutions}, formalized as a \gls{ctmc}.
To make the learning process tractable, they introduce an expanded auxilliary state space that aligns sequences, simultaneously reducing the complexity of marginalizing over possible transitions and enabling efficient element-wise parameterization in sequence space.

In this paper, we unify these perspectives and propose \ours, an Edit Flow for \glspl{tpp} that learns to transport noise sequences $\vt_0\sim p_{\text{noise}}(\vt)$ to data sequences $\vt_1\sim q_{\text{target}}(\vt)$ via atomic \emph{edit operations} insertions, deletions, and substitutions (see \figref{fig:figure1}).
We define these operations specifically for \glspl{tpp}, efficiently parameterize their instantaneous rates within a \gls{ctmc}, propose an auxiliary alignment space for \glspl{tpp}, and show that our unconditionally trained model can be flexibly applied to both unconditional and conditional tasks with adaptive complexity. Our main contributions are:
\begin{itemize}
	\item We introduce \ours, the first generative framework that models \glspl{tpp} via continuous-time edit operations, unifying stochastic set interpolation methods for \glspl{tpp} with Edit Flows for discrete sequences.
	\item We propose a tractable parameterization of insertion, deletion, and substitution rates for \glspl{tpp} within the \gls{ctmc} framework, effectively reducing the number of edit operations for generation.
	\item We demonstrate empirically that \ours\ achieves state-of-the-art results in both unconditional and conditional tasks across diverse real-world and synthetic datasets.
\end{itemize}

\section{Background}\label{sec:background}
\subsection{Temporal Point Processes}\label{sec:tpps}
\looseness=-1
\Glspl{tpp} \citep{daley2006introduction,daley2007introduction} are stochastic processes whose realizations are almost surely finite, ordered sets of random events in time.
Let $\vt = \{t^{(i)}\}_{i=1}^n$, with $t^{(i)} \in [0, T]$, denote a realization of $n$ events on a bounded time interval, which can equivalently be represented by the \textit{counting process} $N(t) = \sum_{i=1}^{n} \mathbf{1}\{t^{(i)} \leq t\}$ counting the number of events up to time $t$.
%We assume the \gls{tpp} is simple, i.e., with probability one no two events coincide: $\Pr(\exists\, i\neq j:\ t^{(i)}=t^{(j)})=0$.
A \gls{tpp} is uniquely characterized by its \textit{conditional intensity function} \citep{rasmussen2018lecturenotestemporalpoint}:
\begin{align}
	\lambda^*(t) = \lim\limits_{\Delta t\downarrow 0} \frac{\mathbb{E}[N(t+\Delta t)-N(t) \mid \mathcal{H}_{t}]}{\Delta t},
\end{align}
where $\mathcal{H}_{t} = \{t^{(i)} : t^{(i)} < t\}$ denotes the history up to time \( t \).
Intuitively, $\lambda^*(t)$ represents the instantaneous rate of events given the past.
Two important properties of \glspl{tpp} are superposition and thinning.
Superposition, i.e., \textit{inserting} one sequence into another, $\vt = \vt_1 \cup \vt_2$, where $\vt_1$ and $\vt_2$ are realizations from \glspl{tpp} with intensities $\lambda_1$ and $\lambda_2$, results in a sample from a \gls{tpp} with intensity $\lambda = \lambda_1 + \lambda_2$.
Independent thinning, i.e., randomly \textit{deleting} any event of a sequence from a \gls{tpp} with intensity $\lambda$ with probability $p$, results in an event sequence from a \gls{tpp} with intensity $(1-p)\lambda$.

The likelihood of observing an event sequence $\vt$ given the conditional intensity/density is:
\begin{equation}
	p(\vt) = \left(\prod_{i=1}^n p(t^{(i)} \mid \mathcal{H}_{t^{(i)}})\right)  \left(1-F(T \mid \mathcal{H}_{t^{(i)}})\right) = \left( \prod_{i=1}^n \lambda^*(t^{(i)}) \right)\text{ exp } \left(-\int_0^T \lambda^*(s) \mathrm{d}s\right),
\end{equation}
where $F(T \mid \mathcal{H}_{t})$ is the CDF of the conditional event density $p(t \mid \mathcal{H}_{t})$.
While this autoregressive formulation of \glspl{tpp} provides a natural framework for modeling event dependencies, it also poses challenges.
Parameterizing the conditional intensity or density is generally nontrivial, and the inherently sequential factorization can lead to inefficient sampling, error accumulation, and limits conditional tasks to forecasting \citep{addthin,psdiff}.

\subsection{Modeling TPPs by set interpolation}
Instead of explicitly modeling the intensity function, \cite{addthin,psdiff} leverage the thinning and superposition properties of \glspl{tpp} to derive diffusion-like generative models that interpolate between data event sequences $\vt_1\sim q_{\text{target}}(\vt)$ and noise $\vt_0\sim p_{\text{noise}}(\vt)$ by inserting and deleting elements.
\addthin\ \citep{addthin} defines the noising Markov chain recursively over a fixed number of steps with size $\Delta$ indexed by $s \in [0,1]$ as follows:
\begin{equation}\label{addthin}
	\lambda_{s}(t) =
	\underbrace{\alpha_{s} \lambda_{s-\Delta}(t)}_{\text{(i) Thin}} +
	\underbrace{(1 - \alpha_s) \lambda_{\mathrm{0}}(t)
	}_{\text{(ii) Add}},
\end{equation}
where $\lambda_{1}(t)$ is the unknown target intensity of the \gls{tpp} and $\alpha_s \in (0,1)$.
Intuitively, this noising process increasingly deletes events from the data sequence, while inserting events from a noise \gls{tpp} $\lambda_{\mathrm{0}}(t)$.
\psdiff\ \citep{psdiff} further separates the adding and thinning to yield a Markov chain for the forward process, that stochastically interpolates between $\vt_0$ and $\vt_1$ as follows:
\begin{equation}\label{eq:conditional}
	p_s(\vt \mid \vt_1, \vt_0) = \prod_{t\in \vt} \begin{cases} \bar{\alpha}_s     & \text{if } t \in \vt_1 \\
              1 - \bar{\alpha}_s & \text{if } t \in \vt_0
	\end{cases}
\end{equation}
or equivalently $\lambda_{s}(t) = \bar{\alpha}_{s} \lambda_{1}(t) + (1 - \bar{\alpha}_s) \lambda_{0}(t)$, with $\bar{\alpha}_{s}$ being the product of $\alpha_i$'s.
\cref{eq:conditional} defines an element-wise conditional path by independent insert and delete operations on \glspl{tpp}, assuming $\vt_0\cap\vt_1 = \emptyset$.
%Since each insertion and deletion is independent, the diffusion posterior $q(\vt_{s+\Delta} \mid \vt_{1}, \vt_s)$ that transports $\vt_0$ to $\vt_1$ can be derived independently per potential set element of $\vt_s$.
%Substitutions on the other hand would couple two elements in $\vt_{1}$ and $\vt_{0}$ consequently breaking the independence of the posterior and \cref{eq:conditional}.
% is parameterized by predicting $\vt_1 \mid \vt_s$ at each $s$.

%However, the recursive definition in \cref{addthin} leads to events from the noise distribution to be added and subsequently removed again.
%They show that predicting $\vt_1 \mid \vt_s$ boils down to classifying the retained target events $\vt_s \cap \vt_1$ and predicting the missing one $\vt_1 \setminus \vt_s$ using an unconditional intensity function based on a Gaussian mixture.

\subsection{Flow matching with edit operations}
\looseness=-1
\citet{editflow} introduce Edit Flows, a non-autoregressive generative framework for variable-length token sequences with a fixed, discrete vocabulary (e.g., language).
They propose a discrete flow that transports a noisy sequence $\vx_0 \sim \pnoise(\vx)$ to a data sequence $\vx_1 \sim \pdata(\vx)$ via elementary \textit{edit operations}: insertions, deletions, and substitutions.
This is formalized via the discrete flow matching framework \citep{gat2024discreteflowmatching,campbell2024generativeflowsdiscretestatespaces,shi2025simplifiedgeneralizedmaskeddiffusion} in an augmented space, yielding a \gls{ctmc} $\Pr(X_{s+h}=\vx \mid X_{s}=\vx_s)=\delta_{\vx_s}(\vx) + h u_s^{\theta}(\vx\mid\vx_s) + o(h)$ with transition rates $u_s^{\theta}$ governed by the edit operations.

Directly defining a conditional rate $u_s(\vx | \vx_1, \vx_0)$ to match $u_s^{\theta}$ to, as in discrete flow matching, is very hard or even intractable, since all possible edits producing $\vx$ must be considered.
Thus, to train this \gls{ctmc}, they rely on two major insights.
First, a \gls{ctmc} in a data space $\dataspace$ can be learned by introducing an augmented space $\dataspace \times \auxspace$ where the true dynamics are known.
Second, designing the auxiliary space $\auxspace$ to follow the element wise mixture probability path $p_{s}(\vz \mid \vz_0, \vz_1) = \prod_n \big[(1-\kappa_s) \delta_{z_0^{(i)}}(z^{(i)}) + \kappa_s \delta_{z_1^{(i)}}(z^{(i)})\big]$ with kappa schedule $\kappa_s \in [0,1]$ \citep{gat2024discreteflowmatching} enables training the \gls{ctmc} directly in the data space $\dataspace$ of variable-length sequences.

Edit operations are encoded by introducing a blank token $\epsilon$ and mapping $(\vx_0, \vx_1)$ into aligned sequences $(\vz_0, \vz_1)$ in $\auxspace$, where pairs \smash{$(z_0^{(i)}, z_1^{(i)})$} correspond to insertions $(\epsilon, x)$, deletions $(x, \epsilon)$, or substitutions $(x, y)$.
Crucially, since the discrete flow matching dynamics in $\auxspace$ are known, they can be transferred back to $\dataspace$ via $p_s(\vx, \vz \mid \vz_0, \vz_1) = p_s(\vz \mid \vz_0, \vz_1)\delta_{\rmblanks(\vz)}(\vx)$, by removing $\epsilon$'s with $\rmblanks$.
Then, the marginal rates $\model$ are learned in $\dataspace$ by marginalizing over $\vz$ with the Bregman divergence
\begin{equation}
	\mathcal{L} = \E_{\substack{(\vz_0, \vz_1)\sim \pi(\vz_0, \vz_1)\\ s, p_{s}(\vz_s, \vx_s  \mid \vz_0, \vz_1) }} \bigg[ \sum_{\vx \neq \vx_s} \model(\vx\mid\vx_s) - \sum_{z_s^{(i)} \neq z_1^{(i)}} \frac{\dot{\kappa}_s}{1-\kappa_s} \log \model\big(\vx(\vz_s, i, z_1^{(i)}) \mid \vx_s \big) \bigg], \label{eq:bregman-divergence}
\end{equation}
where $\vx(\vz_s, i, z_1^{(i)}) = \rmblanks((z_s^{(1)}, \dots, z_s^{(i-1)}, z_1^{(i)}, z_s^{(i+1)}, \dots, z_s^{(n)}))$.

%To support these operations, which modify the sequence length and thus data dimensionality during generation, they rely on two major insights. First, one can learn a \gls{ctmc} in a data space $\dataspace$ based on an augmented space $\dataspace \times \auxspace$ where the true dynamics are known and, second, a clever construction for an auxiliary space $\auxspace$ that introduces a special $\epsilon$ token to uniquely encode edit operations while keeping the data dimensionality constant.

\section{Method}\label{sec:method}

We introduce \ours, an Edit Flow process for \glspl{tpp} that directly learns the joint distribution of event times.
Our process leverages the three elementary edit operations \emph{insert}, \emph{substitute}, and \emph{delete} to define a \gls{ctmc} that continuously interpolates between two event sequences $\vt_0 \sim \pnoise(\vt)$ and $\vt_1 \sim \pdata(\vt)$.

Let $\tppsupport = [0, T]$ denote the support of the \gls{tpp}.
We define the state space as \smash{$\statespace\;=\; \bigcup_{n=0}^{\infty}\left\{ (0, t^{(1)},\dots,t^{(n)}, T) \in \mathcal{T}^n : 0 < t^{(1)} < \cdots < t^{(n)} < T \right\}$}, denoting the set of all possible \emph{padded} \gls{tpp} sequences with finitely many events. Note that the padding values are introduced for notational simplicity when defining the edit operations on $\tppsupport$.

\subsection{Edit operations}\label{sec:edit-operations}
\begin{figure}
	\centering
	\includegraphics[width=0.85\linewidth, trim={5.8cm 7.3cm 7.5cm 5.cm}, clip]{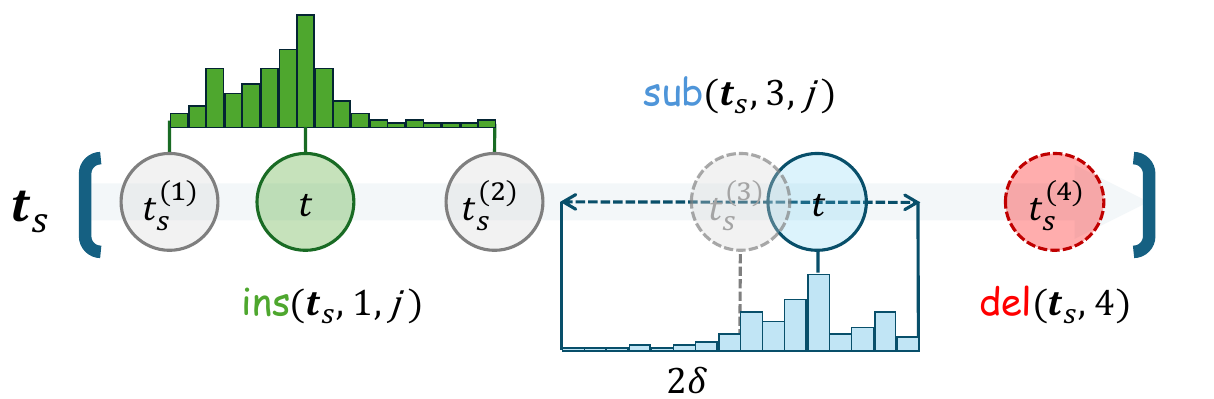}
	\caption{Our discrete edit operations transform continuous event sequences through insertions, substitutions and deletion.}\label{fig:edit-operations}
\end{figure}

Our model navigates the state space $\statespace$ through a set of atomic edit operations.
While Edit Flow was originally defined for discrete state spaces, we can generalize the method to continuous state spaces provided that the set of edit operations remains discrete.
We achieve this by defining a finite set of edit operations on our continuous state space $\statespace$ that nonetheless allow us to transition from any sequence $\vt$ to any other $\vt'$ through repeated application.

Similar to \citet{editflow}, we design our operations to be mutually exclusive: if two sequences differ by exactly one edit, the responsible operation is uniquely determined.
This simplifies the parameterization of the model and computation of the Bregman divergence in \cref{eq:bregman-divergence}.

\textbf{Insertion}: To discretize the event insertion, we quantize the space between any two adjacent events $t^{(i)}$ and $t^{(i+1)}$ into $\insbins$ evenly-spaced bins.
Then, we define the insertion operation relative to the $i$th event as
\begin{equation}\label{psdiff}
	\ins(\vt, i, j) = \Big(t^{(0)}, \dots, t^{(i)}, t^{(i)} + \frac{j - 1 + \alpha}{\insbins} (t^{(i+1)} - t^{(i)}),  t^{(i+1)}, \dots, t^{(n+1)}\Big)
\end{equation}
for $i \in \{0, \dots, n\}$, $j \in [\insbins]$, where $\alpha \sim \uniform(0, 1)$ is a dequantization factor inspired by uniform dequantization in likelihood-based generative models \citep{theis2016note}.
The boundary elements $t^{(0)} = 0$ and $t^{(n+1)} = T$ ensure that insertions are possible across the entire support $\tppsupport$.
Since the bins between different $i$ are non-overlapping, insertions are mutually exclusive.

\textbf{Substitution}: We implement event substitutions by discretizing the continuous space around each event into $\subbins$ bins.
In this case, the bins are free to overlap, since a substitution is always uniquely determined by the substituted event.
We choose a maximum movement distance $\delta$ and define
\begin{equation}
	\sub(\vt, i, j) = \operatorname{sort}\Big(\{t^{(0)}, \dots, t^{(i-1)}, t^{(i+1)}, \dots, t^{(n+1)}\} \cup \Big\{\tilde{t}^{(i)} \Big\}\Big)
\end{equation}
for $i \in \{1, \dots, n\}$, $j \in [\subbins]$, where $\tilde{t}^{(i)} = \big[t^{(i)} - \delta + \frac{j - 1 + \alpha}{\subbins} 2\delta\big]_{0}^{T}$ is the updated event restricted to the support $\tppsupport$ and, again, $\alpha \sim \uniform(0, 1)$ is a uniform dequantization factor within the $j$-th bin.

\textbf{Deletion}: Finally, we define removing event $i \in \{1, \dots, n\}$ straightforwardly as
\begin{equation}
	\del(\vt, i) = (t^{(0)}, \dots, t^{(i-1)}, t^{(i+1)}, \dots, t^{(n+1)}).
\end{equation}
In combination, these operations facilitate any possible edit of an event sequence through insertions and deletions with substitutions as a shortcut for local delete-insert pairs.
Note that we neither allow inserting after the last boundary event nor substituting or deleting the first or last boundary events, thus guaranteeing operations to stay in the state space $\statespace$.
We illustrate the edit operations in \cref{fig:edit-operations}.

Our choice of $\ins$, $\sub$ and $\del$ ensures three key properties: (i) the resulting event sequences remain valid \glspl{tpp}, (ii) the number of valid operations, e.g.\ $\ins(\vt, i, j)$, is independent of the position $i$, which is necessary for efficient parameterization, and (iii) at most one unique operation can transition between any two states, which significantly reduces the complexity of the training loss in \cref{sec:training}.
While these properties are comparably simple to achieve for token sequences in language modeling \citep{editflow}, where any token can replace any other, they require special care in the case of \glspl{tpp}.
$\del$ and $\sub$ are defined to ensure that the resulting event sequence remains in increasing order and that the padding events $t^{(0)}$ and $t^{(n+1)}$ remain in place.
$\del$ transitions are unique because the removed event determines exactly which deletion occurred.
Similarly, $\sub$ transitions are unique because the original position of the substituted event disambiguates the operation, even though two distinct $\sub$ operations may yield the same substituted event value.
To achieve uniqueness for $\ins$, the insertion bins corresponding to $\ins(\vt, i, j)$ have to be mutually disjoint for any $i, j$ since insertions lack a removed event to disambiguate them.
We achieve this by sizing the bins relative to the distance between $t^{(i)}$ and $t^{(i+1)}$.

\paragraph{Parameterization}
\looseness=-1
Generating a new event sequence in the Edit Flow framework then means to emit a continuous stream of edit operations by integrating a rate model $\model(\cdot \mid \vt)$ from $s = 0$ to $s = 1$.
The emitted operations transform a noise sequence $\vt_0$ into a data sample $\vt_1$ by transitioning through a series of intermediate states $\vt$.
Given a current state $\vt_s$, we parameterize the transition rates as
\begin{align}
	\model(\ins(\vt_s, i, j) \mid \vt_s) & = \lambda_{s,i}^{\text{ins}}(\vt_s)\, Q_{s,i}^{\text{ins}}(j \mid \vt_s), \\
	\model(\sub(\vt_s, i, j) \mid \vt_s) & = \lambda_{s,i}^{\text{sub}}(\vt_s)\, Q_{s,i}^{\text{sub}}(j \mid \vt_s), \\
	\model(\del(\vt_s, i) \mid \vt_s)    & = \lambda_{s,i}^{\text{del}}(\vt_s),
\end{align}
where $\lambda_{s,i}^{\text{del}}, \lambda_{s,i}^{\text{ins}}, \lambda_{s,i}^{\text{sub}}$ denote the total rate of each of the three basic operations at each event $t^{(i)}$.
The distributions $Q_{s,i}^{\text{ins}}$ and $Q_{s,i}^{\text{sub}}$ are \emph{categorical} distributions over the discretization bins $j \in [\insbins]$ and $j \in [\subbins]$, respectively.
They distribute the total insertion and substitution rates between the specific options.

%The transition rate $u_s^{\theta}(\vt \mid \vt_s)$ is only non-zero for \emph{neighboring} event sequences $\vt, \vt_s \in \Omega_{\mathcal{T}}$ that differ by exactly one such edit operation.
\subsection{Auxiliary alignment space}\label{sec:alignment}
Training our rate model $\model$ by directly matching a marginalized conditional rate $u_s(\vt \mid \vt_1, \vt_0)$ generating a $p_s(\vt \mid \vt_1, \vt_0)$, as is common in discrete flow matching \citep{campbell2024generativeflowsdiscretestatespaces,gat2024discreteflowmatching},
is challenging or even intractable for Edit Flows, since it would require accounting for all possible edits that could produce $\vt$ \citep{editflow}.

\begin{wrapfigure}[11]{r}{0.3\textwidth}
	\centering
	\vspace{-0.3cm}
	\includegraphics[width=\linewidth, trim={4.6cm 8.3cm 17.cm .8cm},clip]{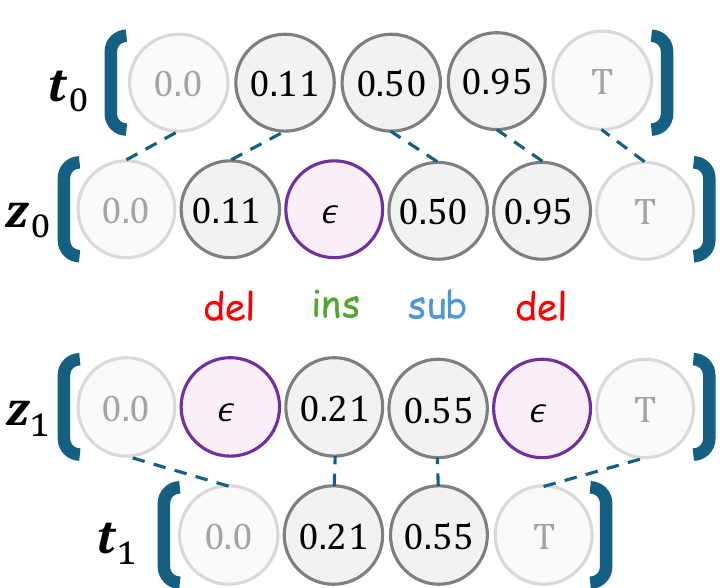}
	\caption{Illustration of the alignment space for $\vt_0$ and $\vt_1$.}\label{fig:figure3}
	\vspace{-1.9cm}
\end{wrapfigure}
To address this, following \cite{editflow}, we introduce an auxiliary alignment space for \glspl{tpp}, where every possible edit operation is uniquely defined in the element wise mixture path $\vz_s \sim p_{s}(\vz_s \mid \vz_0, \vz_1)$, making the learning problem tractable.

In language modeling, any token can appear in any position, so \citet{editflow} achieve strong results even when training with a simple alignment that juxtaposes two sequences after shifting one of them by a constant number of places.
In our case, for the alignments to correspond to possible edit operations, two events can only be matched, i.e., \smash{$z_0^{(i)} \ne \epsilon$} and \smash{$z_1^{(i)} \ne \epsilon$, if $|z_0^{(i)} - z_1^{(i)}| < \delta$} since otherwise the resulting $\sub$ operation would be invalid.
Furthermore, $\vz$s have to correspond to sequences in $\statespace$, so $\rmblanks(\vz)$ has to be increasing, and in particular any mixing $\vz_s$ between $\vz_0$ and $\vz_1$ needs to be valid, i.e., $\vz_s \sim p_{s}(\vz_s \mid \vz_0, \vz_1) \Rightarrow \rmblanks(\vz_s) \in \statespace$.

We find the minimum-cost alignment between the non-boundary events of $\vt_0$ and $\vt_1$ with the Needleman-Wunsch algorithm \citep{needleman1970general}, i.e.,
\begin{equation}
	\alignseqs(\vt_0, \vt_1) = \operatorname{wrap-boundaries}\!\bigg(\!\operatorname{Needleman-Wunsch}\!\Big(\vt_0^{(1:n)}, \vt_1^{(1:m)}, c_{\text{ins}}, c_{\text{sub}}, c_{\text{del}}\Big)\!\bigg)
\end{equation}
and the cost functions
\begin{equation}
	\begin{gathered}
		c_{\text{sub}}(i, j) = \begin{cases}
			|t_0^{(i)} - t_1^{(j)}| & \text{if }|t_0^{(i)} - t_1^{(j)}| < \delta\text{ and }t_0^{(i-1)} < t_1^{(j)} < t_0^{(i+1)} \\
			\infty                  & \text{otherwise}
		\end{cases}\\
		c_{\text{ins}}(i, j) = \begin{cases}
			\frac{\delta}{2} & \text{if }t_0^{(i)} < t_1^{(j)} \\
			\infty           & \text{otherwise}
		\end{cases}
		\qquad
		c_{\text{del}}(i, j) = \begin{cases}
			\frac{\delta}{2} & \text{if }t_0^{(i)} > t_1^{(j)} \\
			\infty           & \text{otherwise}
		\end{cases}
	\end{gathered}\label{eq:nw-costs}
\end{equation}
where $\operatorname{wrap-boundaries}$ wraps the sequences with aligned boundary events $0$ and $T$.
The algorithm builds up the aligned sequences pair by pair.
The operations corresponds to adding different pairs to the end of $(\vz_0, \vz_1)$, i.e., insertion  \smash{$\big(\epsilon, t_1^{(j)}\big)$}, deletion \smash{$\big(t_0^{(i)}, \epsilon\big)$} and substitution \smash{$\big(t_0^{(i)}, t_1^{(j)}\big)$} (see \cref{fig:figure3}).

We carefully craft the cost functions in \cref{eq:nw-costs}, to guarantee that the minimum-cost alignment corresponds to $\ins$, $\sub$ and $\del$ operations as we define them in \cref{sec:edit-operations}.
With the $|t_0^{(i)} - t_1^{(j)}| < \delta$ condition in $c_{\text{sub}}$, we ensure that the aligned sequences will never encode a $\sub$ operation for two events that are further than $\delta$ apart.
The costs for insertions and deletions and the additional condition on $c_{\text{sub}}$ ensure that the aligned sequences are jointly sorted, i.e., for any $i < j$ we have \smash{$\max\big(z_0^{(i)}, z_1^{(i)}\big) < \min\big(z_0^{(j)}, z_1^{(j)}\big)$} where $\min$ and $\max$ ignore $\epsilon$ tokens.
This means that any interpolated $\vz_s$ is sorted by construction.
The validity of encoded $\ins$ and $\del$ operations follows immediately.

%Alignment can be computed in amortized linear time by exploiting the structure of the cost matrix if we assume that the number of elements within $\delta$ of any other element is limited by some constant.

\subsection{Training}\label{sec:training}
We train our model $\model(\cdot \mid \vt_s)$ by optimizing the Bregman divergence in \cref{eq:bregman-divergence}.
This amounts to sampling from a coupling $\pi(\vz_0, \vz_1)$ in the aligned auxiliary space and then matching the ground-truth conditional event rates.
Note that the coupling $\pi(\vz_0, \vz_1)$ is implicitly defined by its sampling procedure: sample $\vt_0, \vt_1 \sim \pi(\vt_0, \vt_1)$ from a coupling of the noise and data distribution, e.g., the independent coupling $\pi(\vt_0, \vt_1) = p(\vt_0)\,q(\vt_1)$, and then align the sequences $\vz_0, \vz_1 = \alignseqs(\vt_0, \vt_1)$.
For our choice of operations, the divergence is
\begin{equation}
	\mathcal{L} = \E_{\substack{(\vz_0, \vz_1)\sim \pi(\vz_0, \vz_1)\\ s, p_{s}(\vz_s, \vt_s  \mid \vz_0, \vz_1) }} \bigg[ \sum_{\operation \in \operations(\vt_s)} \model(\operation \mid \vt_s) - \sum_{z_s^{(i)} \neq z_1^{(i)}} \frac{\dot{\kappa}_s}{1-\kappa_s} \log \model\Big(\operation\big(z_s^{(i)}, z_1^{(i)}\big) \mid \vt_s \Big) \bigg], \label{eq:our-bregman-divergence}
\end{equation}
where $\operations(\vt_s)$ is the set of all edit operations applicable to $\vt_s$ and $\operation\big(z_s^{(i)}, z_1^{(i)}\big)$ is the edit operation encoded in the $i$-th position of the aligned sequences $\vz_s$ and $\vz_1$.
To make it precise, we have
\begin{equation}
	\operations(\vt_s) = \bigcup \begin{cases}
		\{\ins(\vt_s, i, j) \mid i \in \{0\}\,\cup\,[n], j \in [\insbins] \} \\
		\{\sub(\vt_s, i, j) \mid i \in [n], j \in [\subbins] \}              \\
		\{\del(\vt_s, i) \mid i \in [n] \}
	\end{cases}
\end{equation}
and
\begin{equation}
	\operation\big(z_s^{(i)}, z_1^{(i)}\big) = \begin{cases}
		\ins(\vt_s, i', j') & \text{if }z_s^{(i)} = \epsilon\text{ and }z_1^{(i)} \ne \epsilon,   \\
		\sub(\vt_s, i', j') & \text{if }z_s^{(i)} \ne \epsilon\text{ and }z_1^{(i)} \ne \epsilon, \\
		\del(\vt_s, i')     & \text{if }z_s^{(i)} \ne \epsilon\text{ and }z_1^{(i)} = \epsilon.
	\end{cases}
\end{equation}
$i'$ is the index such that $\rmblanks(\vz_s)$ maps $z_s^{(i)}$ to $x_s^{(i')}$ with the convention that $\epsilon$ is mapped to the same $i'$ as the last element of $\vz_s$ before $i$ that is not $\epsilon$.
$j'$ is the index of the insertion or substitution bin relative to $x_s^{(i')}$ that $z_1^{(i)}$ falls into.

\subsection{Sampling}\label{sec:sampling}
\begin{wrapfigure}{r}{0.55\textwidth}
	\vspace{-.9cm}
	\begin{algorithm}[H]
		\caption{Conditional Sampling}\label{alg:sampling}
		\textbf{Input:} \smash{$\text{condition } \vt_1^c = C(\vt_1), \text{ noise } \vt_0 \sim p_{\text{noise}}, h = 1/n_{\text{steps}}$}\;
		\vspace{-0.2cm}
		\smash{$(\vz_0^c, \vz_1^c) \gets \alignseqs(C(\vt_0), \vt_1^c)$}\;
		\vspace{0.2cm}
		\While{$s < 1$}{
		\vspace{0.1cm}
		\textbf{Euler update}\;
		\smash{Sample edits $\operation_s \sim h\, u_s^{\theta}(\cdot \mid \vt_s)$}\;
		\smash{$\vt_{s+h} \gets \text{apply }\operation_s\text{ to }\vt_s$}\;
		\vspace{0.2cm}
		\textbf{Recondition}\;
		\smash{$\tilde{\vz}_{s+h}^c \sim p_{s+h}(\cdot \mid \vz_0^c, \vz_1^c)$}\;
		\smash{$\vt_{s+h}^c \gets \rmblanks(\tilde{\vz}_{s+h}^c)$}\;
		\vspace{0.2cm}
		\textbf{Merge}\;
		\smash{$\vt_{s+h} \gets C'(\vt_{s+h}) \cup \vt_{s+h}^c$}\;
		\vspace{0.2cm}
		\smash{$s \gets s + h$}\;
		}
		\textbf{Return:} forecast trajectory $C'(\vt_{s=1})$
	\end{algorithm}
	\vspace{-.8cm}
\end{wrapfigure}
Sampling from our model is done by forward simulation of the \gls{ctmc} from noise $\vt_0 \sim p_{\text{noise}}(\vt)$ up to $s=1$.
We follow \citep{editflow,gat2024discreteflowmatching} and leverage their Euler approximation, since exact simulation is intractable.
Even though the rates are parameterized per element, sampling multiple edits within a time horizon can be done in parallel.
At each step of length $h$, insertions at position $i$ occur with probability \smash{$h\,\lambda_{s,i}^{\text{ins}}(\vt)$} and deletions or substitutions occur with probability \smash{$h(\lambda_{s,i}^{\text{del}}(\vt) + \lambda_{s,i}^{\text{sub}}(\vt))$}.
Since they are mutually exclusive the probability of substitution vs deletion is $\lambda_{s,i}^{\text{sub}}(\vt)/ (\lambda_{s,i}^{\text{sub}}(\vt) + \lambda_{s,i}^{\text{del}}(\vt))$.
Lastly, the inserted or substituted events are drawn from the respective distributions $Q$ to update $\vt_s$.
For a short summary of the unconditional sampling step refer to the Euler update step depicted in algorithm \cref{alg:sampling}.

\textbf{Conditional sampling.} We can extend the unconditional model to conditional generation given a binary mask on time $c: \mathcal{T} \rightarrow \{0,1\}$ (e.g., for forecasting, $c(t) = t \leq t_{\text{history}}$).
For a sequence $\vt$, we define the conditioned part $C(\vt)=\{t \in \vt : c(t)=1\}$ and its complement $C'(\vt)$.
Then as depicted in algorithm \cref{alg:sampling}, for conditional sampling, we can simply enforce the conditional subsequence to follow a noisy interpolation between $\vt_0^c=C(\vt_0)$ and $\vt_1^c=C(\vt_1)$, while the complement evolves freely in the sampling process.

\subsection{Model Architecture}\label{sec:model-architecture}

For our rate model $\model(\cdot \mid \vx_s)$, we adapt the Llama architecture, a transformer widely applied for variable-length sequences in language modeling \citep{touvron2023llama}.
We employ \texttt{FlexAttention} in the Llama attention blocks, which supports variable-length sequences natively without padding \citep{dong2024flex}.
As a first step, we convert the scalar event sequence $\vx_s$ into a sequence of token embeddings by applying \smash{$\operatorname{MLP}(\operatorname{SinEmb}(x_s^{(i)} / T))$} to each to each event, where $\operatorname{MLP}$ refers to a small \gls{mlp} and $\operatorname{SinEmb}$ is a sinusoidal embedding \citep{vaswani2017attention}.
We convert $s$ and $|\vx_s|$ into two additional tokens in an equivalent way with separate \glspl{mlp} and prepend them to the embedding sequence, which we then feed to the Llama.
Lastly, we apply one more \gls{mlp} to map the output embedding $\vh^{(i)}$ of each event to transition rates.
In particular, we parameterize
\begin{gather}
	\lambda_{s,i}^{\text{ins}} = \exp(\lambda_{\text{M}}\tanh(\vh_{\text{ins}}^{(i)})), \quad \lambda_{s,i}^{\text{sub}} = \exp(\lambda_{\text{M}}\tanh(\vh_{\text{sub}}^{(i)})), \quad \lambda_{s,i}^{\text{del}} = \exp(\lambda_{\text{M}}\tanh(\vh_{\text{del}}^{(i)})), \\
	\mQ_{s,i}^{\text{ins}} = \operatorname{softmax}(\vh_{\mQ,\text{ins}}^{(i)}), \quad \mQ_{s,i}^{\text{sub}} = \operatorname{softmax}(\vh_{\mQ,\text{sub}}^{(i)}).
\end{gather}
We list the values of all relevant hyperparameters in \cref{sec:model-parameters}.

\section{Experiments}\label{sec:experiments}
We evaluate our model on seven real-world and six synthetic benchmark datasets \citep{omi2019fully,shchur2020fast,addthin,psdiff}.
In our experiments, we compare against \iftpp\ \citep{shchur2019intensity}, an autoregressive baseline which consistently shows state-of-the-art performance \citep{bosser2023on,addthin,eventflow}.
We further compare to \psdiff\ \citep{psdiff} and \addthin\ \citep{addthin}, given their strong results in both conditional and unconditional settings and their methodological similarity to our approach.
All models are trained with five seeds and we select the best checkpoint based on $W_1$-over-$\diet$ against a validation set.
\ours, \addthin, and \psdiff\ are trained unconditionally but can be conditioned at inference time.\footnote{To stay comparable, we employ the conditioning algorithm from \cite{psdiff} for \addthin.}
We list the full results in \cref{sec:tables}.

For forecasts, we compare predicted and target sequences by three metrics:
$\dxiao$ introduced by \citet{xiao2017wasserstein}, the \gls{mre} of the event counts and $\diet$, which compares inter-event times to quantify the relation between events such as burstiness.
In unconditional generation, we compare our generated sequences to the test set in terms of \gls{mmd} \citep{shchur2020fast} and their Wasserstein-1 distance with respect to their counts ($\dlen$) and inter-event times ($\diet$).
See \cref{sec:metrics} for details.

\subsection{Unconditional generation}
\begin{table}[t]
	\centering
	\caption{Unconditional sampling performance. Bold is best, underlined second best. Ranking follows full results in \cref{sec:tables} and results are grouped if they fall within the std of the best member.}\label{tab:conditional}
	\begin{tabular}{rlr@{\hspace{5pt}}r@{\hspace{5pt}}r@{\hspace{5pt}}r@{\hspace{5pt}}r@{\hspace{5pt}}r@{\hspace{5pt}}r@{\hspace{5pt}}r@{\hspace{5pt}}r@{\hspace{5pt}}r@{\hspace{5pt}}r@{\hspace{5pt}}r@{\hspace{5pt}}r@{\hspace{5pt}}}
\toprule
{} & {} & {H1} & {H2} & {NSP} & {NSR} & {SC} & {SR} & {PG} & {R/C} & {R/P} & {Tx} & {Tw} & {Y/A} & {Y/M} \\
\midrule
\multirow[c]{5}{*}{\rotatebox[origin=c]{90}{$\operatorname{MMD}$}} & \iftpp & \underline{\num{1.6}} & \textbf{\num{1.2}} & \underline{\num{3.2}} & \underline{\num{3.9}} & \textbf{\num{6.7}} & \textbf{\num{1.2}} & \num{16.2} & \textbf{\num{7.5}} & \underline{\num{2.0}} & \num{5.0} & \underline{\num{2.6}} & \num{5.8} & \textbf{\num{2.9}} \\
 & \addthin & \num{2.4} & \underline{\num{1.8}} & \underline{\num{3.5}} & \num{15.7} & \num{24.6} & \underline{\num{2.5}} & \num{4.6} & \underline{\num{63.0}} & \num{10.2} & \underline{\num{4.1}} & \num{4.4} & \num{11.8} & \underline{\num{3.7}} \\
 & \psdiff & \num{3.3} & \underline{\num{1.8}} & \textbf{\num{2.0}} & \num{5.9} & \underline{\num{19.8}} & \underline{\num{2.4}} & \underline{\num{3.2}} & \textbf{\num{6.5}} & \textbf{\num{1.0}} & \underline{\num{3.8}} & \num{3.4} & \underline{\num{4.1}} & \underline{\num{3.4}} \\
 & \ours{} & \textbf{\num{1.1}} & \textbf{\num{1.2}} & \textbf{\num{1.7}} & \textbf{\num{3.5}} & \textbf{\num{7.7}} & \textbf{\num{1.0}} & \textbf{\num{1.4}} & \textbf{\num{8.2}} & \underline{\num{2.4}} & \textbf{\num{3.1}} & \textbf{\num{1.3}} & \textbf{\num{3.7}} & \underline{\num{4.0}} \\
 &  & \scalebox{0.5}{$\times 10^{-2}$} & \scalebox{0.5}{$\times 10^{-2}$} & \scalebox{0.5}{$\times 10^{-2}$} & \scalebox{0.5}{$\times 10^{-2}$} & \scalebox{0.5}{$\times 10^{-2}$} & \scalebox{0.5}{$\times 10^{-2}$} & \scalebox{0.5}{$\times 10^{-2}$} & \scalebox{0.5}{$\times 10^{-3}$} & \scalebox{0.5}{$\times 10^{-2}$} & \scalebox{0.5}{$\times 10^{-2}$} & \scalebox{0.5}{$\times 10^{-2}$} & \scalebox{0.5}{$\times 10^{-2}$} & \scalebox{0.5}{$\times 10^{-2}$} \\
\midrule\multirow[c]{5}{*}{\rotatebox[origin=c]{90}{$W_{1,\dlen}$}} & \iftpp & \underline{\num{20.5}} & \underline{\num{13.3}} & \num{11.5} & \num{14.1} & \textbf{\num{1.5}} & \underline{\num{23.0}} & \num{294.6} & \num{3.9} & \underline{\num{3.2}} & \underline{\num{2.9}} & \underline{\num{6.5}} & \underline{\num{3.3}} & \num{2.5} \\
 & \addthin & \num{33.3} & \num{21.8} & \num{12.8} & \num{49.0} & \num{22.7} & \num{41.8} & \underline{\num{24.5}} & \num{37.0} & \num{33.6} & \textbf{\num{2.3}} & \num{15.5} & \num{6.0} & \textbf{\num{1.6}} \\
 & \psdiff & \num{26.9} & \num{29.6} & \underline{\num{5.5}} & \underline{\num{13.3}} & \underline{\num{10.6}} & \num{30.3} & \underline{\num{16.1}} & \textbf{\num{1.3}} & \textbf{\num{2.5}} & \textbf{\num{2.8}} & \underline{\num{6.3}} & \textbf{\num{1.5}} & \textbf{\num{1.5}} \\
 & \ours{} & \textbf{\num{7.6}} & \textbf{\num{7.0}} & \textbf{\num{3.1}} & \textbf{\num{1.5}} & \textbf{\num{1.3}} & \textbf{\num{6.4}} & \textbf{\num{6.2}} & \underline{\num{1.9}} & \num{5.7} & \textbf{\num{2.5}} & \textbf{\num{3.4}} & \textbf{\num{1.4}} & \underline{\num{1.7}} \\
 &  & \scalebox{0.5}{$\times 10^{-3}$} & \scalebox{0.5}{$\times 10^{-3}$} & \scalebox{0.5}{$\times 10^{-3}$} & \scalebox{0.5}{$\times 10^{-3}$} & \scalebox{0.5}{$\times 10^{-3}$} & \scalebox{0.5}{$\times 10^{-3}$} & \scalebox{0.5}{$\times 10^{-3}$} & \scalebox{0.5}{$\times 10^{-2}$} & \scalebox{0.5}{$\times 10^{-2}$} & \scalebox{0.5}{$\times 10^{-2}$} & \scalebox{0.5}{$\times 10^{-3}$} & \scalebox{0.5}{$\times 10^{-2}$} & \scalebox{0.5}{$\times 10^{-2}$} \\
\midrule\multirow[c]{5}{*}{\rotatebox[origin=c]{90}{$W_{1,\diet}$}} & \iftpp & \underline{\num{6.3}} & \underline{\num{5.8}} & \underline{\num{3.2}} & \textbf{\num{2.3}} & \underline{\num{6.5}} & \textbf{\num{7.1}} & \num{30.3} & \num{1.8} & \num{7.1} & \num{17.4} & \underline{\num{4.9}} & \underline{\num{3.2}} & \num{2.8} \\
 & \addthin & \underline{\num{6.6}} & \num{7.0} & \underline{\num{3.2}} & \underline{\num{3.9}} & \num{15.1} & \underline{\num{9.4}} & \underline{\num{8.0}} & \num{5.3} & \num{20.0} & \textbf{\num{8.8}} & \num{5.5} & \underline{\num{3.2}} & \underline{\num{2.4}} \\
 & \psdiff & \num{8.6} & \num{9.9} & \textbf{\num{3.0}} & \num{5.1} & \num{32.6} & \num{12.8} & \num{9.0} & \underline{\num{1.6}} & \textbf{\num{4.3}} & \underline{\num{11.1}} & \num{6.7} & \textbf{\num{2.4}} & \textbf{\num{2.3}} \\
 & \ours{} & \textbf{\num{5.3}} & \textbf{\num{5.5}} & \textbf{\num{3.1}} & \textbf{\num{2.2}} & \textbf{\num{6.4}} & \textbf{\num{7.0}} & \textbf{\num{7.5}} & \textbf{\num{1.4}} & \underline{\num{6.0}} & \underline{\num{11.1}} & \textbf{\num{4.6}} & \textbf{\num{2.5}} & \textbf{\num{2.3}} \\
 &  & \scalebox{0.5}{$\times 10^{-1}$} & \scalebox{0.5}{$\times 10^{-1}$} & \scalebox{0.5}{$\times 10^{-1}$} & \scalebox{0.5}{$\times 10^{-1}$} & \scalebox{0.5}{$\times 10^{-2}$} & \scalebox{0.5}{$\times 10^{-1}$} & \scalebox{0.5}{$\times 10^{-2}$} & \scalebox{0.5}{$\times 10^{-1}$} & \scalebox{0.5}{$\times 10^{-3}$} & \scalebox{0.5}{$\times 10^{-2}$} & \scalebox{0.5}{$\times 10^{-1}$} & \scalebox{0.5}{$\times 10^{-1}$} & \scalebox{0.5}{$\times 10^{-1}$} \\
\bottomrule
\end{tabular}

\end{table}
To evaluate how well samples from each \gls{tpp} model follow the data distribution, we compute distance metrics between 4000 sampled sequences and a hold-out test set.
We report the unconditional sampling results in \cref{tab:conditional}.
\ours\ achieves the best rank in unconditional sampling by strongly matching the test set distribution across all evaluation metrics, outperforming all baselines.
The autoregressive baseline \iftpp\ shows very strong unconditional sampling capability, closely matching and on some dataset and metric combination outperforming the other non-autoregressive baselines \addthin\ and \psdiff.

\subsection{Conditional generation (Forecasting)}

Predicting the future given some history window is a fundamental \gls{tpp} task.
For each test sequence, we uniformly sample 50 forecasting windows $[T_0, T], T_0\in [\Delta T, T-\Delta T]$, with minimal history and forecast time $\Delta T$.
While, this set-up is very similar to the one proposed by \cite{addthin}, there are key differences: we do not fix the forecast window and do not enforce a minimal number of forecast or history events.
In fact, even an empty history encodes the information of not having observed an event and a \gls{tpp} should capture the probability of not observing any event in the future.

\begin{wraptable}{r}{0.6\textwidth}
	\centering
	\caption{Forecasting accuracy up to $T$. Bold is best, underlined second best. Ranking follows full results in \cref{sec:tables} and results are grouped if they fall within the std of the best member.}\label{tab:forecast}
	\begin{tabular}{rlr@{\hspace{5pt}}r@{\hspace{5pt}}r@{\hspace{5pt}}r@{\hspace{5pt}}r@{\hspace{5pt}}r@{\hspace{5pt}}r@{\hspace{5pt}}}
\toprule
{} & {} & {PG} & {R/C} & {R/P} & {Tx} & {Tw} & {Y/A} & {Y/M} \\
\midrule
\multirow[c]{5}{*}{\rotatebox[origin=c]{90}{$\dxiao$}} & \iftpp & \num{6.0} & \num{3.9} & \underline{\num{6.3}} & \num{4.7} & \textbf{\num{2.6}} & \num{1.8} & \num{3.4} \\
 & \addthin & \underline{\num{2.5}} & \num{8.8} & \underline{\num{7.3}} & \textbf{\num{4.0}} & \num{2.8} & \underline{\num{1.5}} & \textbf{\num{2.9}} \\
 & \psdiff & \textbf{\num{2.4}} & \textbf{\num{3.2}} & \textbf{\num{4.8}} & \underline{\num{4.4}} & \underline{\num{2.6}} & \textbf{\num{1.5}} & \underline{\num{3.0}} \\
 & \ours{} & \underline{\num{2.5}} & \underline{\num{3.4}} & \textbf{\num{4.9}} & \underline{\num{4.5}} & \underline{\num{2.7}} & \textbf{\num{1.5}} & \underline{\num{3.0}} \\
 &  &  & \scalebox{0.5}{$\times 10^{1}$} & \scalebox{0.5}{$\times 10^{1}$} &  &  &  &  \\
\midrule\multirow[c]{5}{*}{\rotatebox[origin=c]{90}{MRE}} & \iftpp & \num{38.9} & \num{7.5} & \num{3.5} & \underline{\num{3.2}} & \textbf{\num{2.1}} & \num{3.7} & \underline{\num{3.9}} \\
 & \addthin & \num{3.7} & \num{14.8} & \num{4.6} & \textbf{\num{3.0}} & \num{3.0} & \textbf{\num{3.5}} & \textbf{\num{3.7}} \\
 & \psdiff & \textbf{\num{3.4}} & \textbf{\num{3.3}} & \underline{\num{3.0}} & \num{11.4} & \num{2.4} & \textbf{\num{3.5}} & \num{9.2} \\
 & \ours{} & \underline{\num{3.5}} & \underline{\num{3.6}} & \textbf{\num{2.8}} & \num{12.3} & \underline{\num{2.3}} & \underline{\num{3.5}} & \num{9.0} \\
 &  & \scalebox{0.5}{$\times 10^{-1}$} &  & \scalebox{0.5}{$\times 10^{-1}$} & \scalebox{0.5}{$\times 10^{-1}$} &  & \scalebox{0.5}{$\times 10^{-1}$} & \scalebox{0.5}{$\times 10^{-1}$} \\
\midrule\multirow[c]{5}{*}{\rotatebox[origin=c]{90}{$\diet$}} & \iftpp & \num{4.7} & \underline{\num{6.8}} & \num{14.7} & \num{1.4} & \num{2.2} & \num{5.9} & \num{3.9} \\
 & \addthin & \underline{\num{4.0}} & \underline{\num{6.9}} & \underline{\num{10.3}} & \underline{\num{1.2}} & \underline{\num{1.5}} & \textbf{\num{4.9}} & \textbf{\num{2.6}} \\
 & \psdiff & \underline{\num{4.1}} & \textbf{\num{6.2}} & \textbf{\num{9.5}} & \textbf{\num{1.1}} & \underline{\num{1.5}} & \textbf{\num{4.9}} & \textbf{\num{2.6}} \\
 & \ours{} & \textbf{\num{4.0}} & \underline{\num{6.8}} & \underline{\num{10.1}} & \textbf{\num{1.1}} & \textbf{\num{1.4}} & \underline{\num{5.0}} & \underline{\num{2.7}} \\
 &  & \scalebox{0.5}{$\times 10^{-1}$} & \scalebox{0.5}{$\times 10^{-1}$} & \scalebox{0.5}{$\times 10^{-3}$} & \scalebox{0.5}{$\times 10^{-1}$} &  & \scalebox{0.5}{$\times 10^{-1}$} & \scalebox{0.5}{$\times 10^{-1}$} \\
\bottomrule
\end{tabular}

	\vspace{-0.5cm}
\end{wraptable}
We report the forecasting results in \cref{tab:forecast}.
\ours\ shows very strong forecasting capabilities closely matching or surpassing the baselines across most dataset and metric combinations.
Even though \iftpp\ is explicitly trained to auto-regressively predict the next event given its history, it shows overall worse forecasting capabilities compared to the unconditionally trained \ours, \addthin\ and \psdiff.
This again, underlines previous findings \citep{addthin}, that autoregressive \glspl{tpp} can suffer from error accumulation in forecasting.
Similar to the unconditional setting, \psdiff\ (transformer) outperforms \addthin\ (convolution with circular padding), which showcases the improved posterior and modeling of long-range interactions.

\subsection{Edit efficiency}
The $\sub$ operation allows our model to modify sequences in a more targeted way when compared to \psdiff{} or \addthin{}, which have to rely on just inserts and deletes.
\begin{wraptable}[8]{r}{0.48\textwidth}
	\centering
	\caption{Average number of edit operations in unconditional sampling across datasets. Full result in \cref{tab:edit_ops_full}.}
	\label{tab:edit-ops}
	\small
	\setlength{\tabcolsep}{5pt}
	\renewcommand{\arraystretch}{0.95}
	\begin{tabular}{l|ccc|c}
		\toprule
		        & Ins    & Del   & Sub   & Total  \\
		\midrule
		\psdiff & 173.48 & 61.04 & 0.00  & 234.52 \\
		\ours   & 137.42 & 33.08 & 29.16 & 199.65 \\
		\bottomrule
	\end{tabular}
\end{wraptable}
Note that one $\sub$ operation can replace an insert-delete pair.
\cref{tab:edit-ops} shows this results in \ours{} using fewer edit operations than \psdiff{} on average even if one would count substitutions twice, as an insert and a delete.\footnote{Due to its recursive definition, \addthin\ inserts and subsequently deletes some noise events during sampling, which results in additional edit operations compared to \psdiff.}
This is further amplified by the fact, that unlike \ours, \psdiff\ and \addthin{} only indirectly parameterizes the transition edit rates by predicting $\vt_1$ by insertion and deletion at every sampling step.
\begin{figure*}[t]
	\centering
	\includegraphics{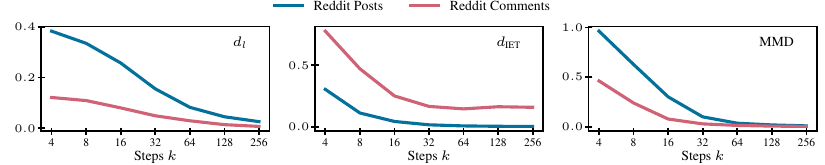}
	\caption{Changing the number of steps $k$ allows trading off compute and sample quality in terms of $\dlen$, $\diet$ and MMD at inference time.}\label{fig:metrics-steps}
\end{figure*}

\begin{wraptable}[8]{r}{0.48\textwidth}
	\centering
	\caption{Sample run-time (ms) on a H100 GPU.}
	\label{tab:sample_runtime}
	\begin{tabular}{lcc}
		\toprule
		         & {R/P}     & {R/C}     \\
		\midrule
		\addthin & 18,075.62 & 17,689.36 \\
		\psdiff  & 7,776.35  & 3,913.78  \\
		\ours    & 4,120.38  & 1,505.68  \\
		\bottomrule
	\end{tabular}
	\vspace{0.cm}
\end{wraptable}
In \cref{tab:sample_runtime}, we compare their actual sampling runtime for a batch size of 1024 on the two dataset with the longest sequences.
Our implementation beats the reference implementations of \addthin{} and \psdiff{} by a large margin.
Note, that for a fair comparison, we fixed the number of sampling steps to 100 in all previous evaluations.
As a continuous-time model, \ours{} can further trade off compute against sample quality at inference time without retraining, in contrast to discrete-time models like \addthin{} and \psdiff{}.
\cref{fig:metrics-steps} shows that sample quality improves as we increase the number of sampling steps and therefore reduce the discretization step size of the \gls{ctmc} dynamics.
At the same time, the figure also shows rapidly diminishing quality improvements, highlighting potential for substantial speedups with only minor quality loss.

% As a consequence, to generate a sequence $\vt_1$ of length $n$, the expected total number of actual insert operations depends on the posterior probability $p_s$ of retaining a predicted event $\vt_1 \mid \vt_s$ and is given by $n \sum_{s=1}^{\text{steps}} \prod_{i=1}^{s} (1 - p_i)$.
% Even for $n=100$, this already amounts to approximately $5{,}050$ insertions under their alpha schedule.
% However, with modern compute these edits can be applied in parallel, so generation runtimes scale primarily with the number of sampling steps.

% Kind of negative on this paragraph. Should rather say more about results tables.
%Lastly, there are technical differences: we parameterize per-position insertion rates, enabling a very efficient discrete parametrization, while \psdiff\ parameterizes insertions continuously over the whole domain, requiring rejection sampling.
%When encoding the current state $\vt_s$, we employ a flex-attention mechanism, which avoids padding and makes efficient use of modern compute.

\section{Related work}\label{sec:related-work}
The statistical modeling of \glspl{tpp} has a long history \citep{daley2007introduction,hawkes1971spectra}.
Classical approaches such as the Hawkes process define parametric conditional intensities, but their limited flexibility has motivated the development of neurally parameterized \glspl{tpp}:

\textbf{Autoregressive Neural TPP}:
Most neural \glspl{tpp} adopt an autoregressive formulation, modeling the distribution of each event conditional on its history.
These models consist of two components: a \emph{history encoder} and an \emph{event decoder}.
\emph{Encoders} are typically implemented using recurrent neural networks \citep{du2016recurrent,shchur2019intensity} or attention mechanisms \citep{zhang2020self,zuo2020transformer,mei2022transformer}, with attention-based models providing longer-range context at the cost of higher complexity \citep{shchur2021neural}.
Further, some propose to encode the history of a \gls{tpp} in a continuous latent stochastic processes \citep{chen2020neural,enguehard2020neural,jia2019neural,hasan2023inference}.
For the \emph{decoder}, a wide variety of parametrizations have been explored.
Conditional intensities or related measures (e.g., hazard function or conditional density), can be modeled, parametrically \citep{neuralhawkes,zuo2020transformer,zhang2020self}, via neural networks \citep{omi2019fully}, mixtures of kernels \citep{okawa2019deep,soen2021unipoint,zhang2020cause} and mixture distributions \citep{shchur2019intensity}.
Generative approaches further enhance flexibility: normalizing flow-based  \citep{shchur2020fast}, GAN-based \citep{xiao2017wasserstein}, VAE-based \citep{li2018learning}, and diffusion-based decoders \citep{linexploring,yuan2023diffstpp} have all been proposed.
While expressive, autoregressive \glspl{tpp} are inherently sequential, which makes sampling scale at least linearly with sequence length, can lead to error accumulation in multi-step forecasting and limit conditional generation to forecasting.

\textbf{Non-autoregressive Neural TPPs}:
Similar to our method, these approaches model event sequences through a latent variable process that refines the entire sequence jointly.
Diffusion-inspired \citep{addthin,psdiff} and flow-based generative models \citep{eventflow} have recently emerged as promising alternatives to auto-regressive TPP models by directly modelling the joint distribution over event sequences.

%One line of work draws inspiration from Cox processes \citep{cox1955some,moller1998lgcp,deep_neyman_scott}. While flexible, these models require approximating intractable integrals, making training and sampling computationally demanding.

\section{Conclusion}\label{sec:conclusion}
We have presented \ours, an Edit Flow for \glspl{tpp} that generalises diffusion-based set interpolation methods \citep{addthin,psdiff} with a continuous-time flow model introducing substitution as an additional edit operation.
By parameterizing insertions, deletions, and substitutions within a \gls{ctmc}, our approach enables efficient and flexible sequence modeling for \glspl{tpp}.
Empirical results demonstrate that \ours\ matches state-of-the-art performance in both unconditional and conditional generation tasks across synthetic and real-world datasets, while reducing the number of edit operations.

\section*{Acknowledgments}\label{sec:acknowledgments}

We want to thank the Munich Center for Machine Learning for providing compute resources.

\newpage

\bibliography{iclr2026_conference}

@book{daley2007introduction,
  title={An introduction to the theory of point processes: volume II: general theory and structure},
  author={Daley, Daryl J and Vere-Jones, David},
  year={2007},
  publisher={Springer Science \& Business Media}
}

@article{hawkes1971spectra,
  title={Spectra of some self-exciting and mutually exciting point processes},
  author={Hawkes, Alan G},
  journal={Biometrika},
  volume={58},
  number={1},
  pages={83--90},
  year={1971},
  publisher={Oxford University Press}
}

@book{daley2006introduction,
  title={An Introduction to the Theory of Point Processes: Volume I: Elementary Theory and Methods},
  author={Daley, D.J. and Vere-Jones, D.},
  series={Probability and Its Applications},
  year={2006},
  publisher={Springer New York}
}

@inproceedings{shchur2019intensity,
  title={Intensity-Free Learning of Temporal Point Processes},
  author={Shchur, Oleksandr and Bilo{\v{s}}, Marin and G{\"u}nnemann, Stephan},
  booktitle={International Conference on Learning Representations (ICLR)},
  year={2020},
}

@inproceedings{du2016recurrent,
  title={Recurrent marked temporal point processes: Embedding event history to vector},
  author={Du, Nan and Dai, Hanjun and Trivedi, Rakshit and Upadhyay, Utkarsh and Gomez-Rodriguez, Manuel and Song, Le},
  booktitle={Proceedings of the 22nd ACM SIGKDD international conference on knowledge discovery and data mining},
  pages={1555--1564},
  year={2016}
}

@inproceedings{neuralhawkes,
  title={The Neural Hawkes Process: A Neurally Self-Modulating Multivariate Point Process},
  author={Mei, Hongyuan and Eisner, Jason M},
  booktitle={Neural Information Processing Systems (NeurIPS)},
  year={2017}
}

@article{zuo2020transformer,
  title={Transformer Hawkes Process},
  author={Zuo, Simiao and Jiang, Haoming and Li, Zichong and Zhao, Tuo and Zha, Hongyuan},
  journal={arXiv preprint arXiv:2002.09291},
  year={2020}
}

@inproceedings{soen2021unipoint,
  title={UNIPoint: Universally Approximating Point Processes Intensities},
  author={Soen, Alexander and Mathews, Alexander and Grixti-Cheng, Daniel and Xie, Lexing},
  booktitle={Proceedings of the AAAI Conference on Artificial Intelligence},
  volume={35},
  pages={9685--9694},
  year={2021}
}

@article{omi2019fully,
  title={Fully neural network based model for general temporal point processes},
  author={Omi, Takahiro and Aihara, Kazuyuki and others},
  journal={Advances in neural information processing systems},
  volume={32},
  year={2019}
}

@inproceedings{
mei2022transformer,
title={Transformer Embeddings of Irregularly Spaced Events and Their Participants},
author={Hongyuan Mei and Chenghao Yang and Jason Eisner},
booktitle={International Conference on Learning Representations},
year={2022},
url={https://openreview.net/forum?id=Rty5g9imm7H}
}

@article{chen2020neural,
  title={Neural spatio-temporal point processes},
  author={Chen, Ricky TQ and Amos, Brandon and Nickel, Maximilian},
  journal={arXiv preprint arXiv:2011.04583},
  year={2020}
}

@inproceedings{enguehard2020neural,
  title={Neural temporal point processes for modelling electronic health records},
  author={Enguehard, Joseph and Busbridge, Dan and Bozson, Adam and Woodcock, Claire and Hammerla, Nils},
  booktitle={Machine Learning for Health},
  pages={85--113},
  year={2020},
  organization={PMLR}
}

@inproceedings{
hasan2023inference,
title={Inference and Sampling of Point Processes from Diffusion Excursions},
author={Ali Hasan and Yu Chen and Yuting Ng and Mohamed Abdelghani and Anderson Schneider and Vahid Tarokh},
booktitle={The 39th Conference on Uncertainty in Artificial Intelligence},
year={2023}
}

@article{jia2019neural,
  title={Neural jump stochastic differential equations},
  author={Jia, Junteng and Benson, Austin R},
  journal={Advances in Neural Information Processing Systems},
  volume={32},
  year={2019}
}

@misc{rasmussen2018lecturenotestemporalpoint,
      title={Lecture Notes: Temporal Point Processes and the Conditional Intensity Function}, 
      author={Jakob Gulddahl Rasmussen},
      year={2018},
      eprint={1806.00221},
      archivePrefix={arXiv},
      primaryClass={stat.ME},
      url={https://arxiv.org/abs/1806.00221}, 
}

@misc{eventflow,
      title={EventFlow: Forecasting Temporal Point Processes with Flow Matching}, 
      author={Gavin Kerrigan and Kai Nelson and Padhraic Smyth},
      year={2025},
      eprint={2410.07430},
      archivePrefix={arXiv},
      primaryClass={cs.LG},
      url={https://arxiv.org/abs/2410.07430}, 
}

@misc{gat2024discreteflowmatching,
      title={Discrete Flow Matching}, 
      author={Itai Gat and Tal Remez and Neta Shaul and Felix Kreuk and Ricky T. Q. Chen and Gabriel Synnaeve and Yossi Adi and Yaron Lipman},
      year={2024},
      eprint={2407.15595},
      archivePrefix={arXiv},
      primaryClass={cs.LG},
      url={https://arxiv.org/abs/2407.15595}, 
}

@article{
bosser2023on,
title={On the Predictive Accuracy of Neural Temporal Point Process Models for Continuous-time Event Data},
author={Tanguy Bosser and Souhaib Ben Taieb},
journal={Transactions on Machine Learning Research},
issn={2835-8856},
year={2023},
url={https://openreview.net/forum?id=3OSISBQPrM},
note={Survey Certification}
}

@misc{shi2025simplifiedgeneralizedmaskeddiffusion,
      title={Simplified and Generalized Masked Diffusion for Discrete Data}, 
      author={Jiaxin Shi and Kehang Han and Zhe Wang and Arnaud Doucet and Michalis K. Titsias},
      year={2025},
      eprint={2406.04329},
      archivePrefix={arXiv},
      primaryClass={cs.LG},
      url={https://arxiv.org/abs/2406.04329}, 
}

@misc{campbell2024generativeflowsdiscretestatespaces,
      title={Generative Flows on Discrete State-Spaces: Enabling Multimodal Flows with Applications to Protein Co-Design}, 
      author={Andrew Campbell and Jason Yim and Regina Barzilay and Tom Rainforth and Tommi Jaakkola},
      year={2024},
      eprint={2402.04997},
      archivePrefix={arXiv},
      primaryClass={stat.ML},
      url={https://arxiv.org/abs/2402.04997}, 
}

@misc{editflow,
      title={Edit Flows: Flow Matching with Edit Operations}, 
      author={Marton Havasi and Brian Karrer and Itai Gat and Ricky T. Q. Chen},
      year={2025},
      eprint={2506.09018},
      archivePrefix={arXiv},
      primaryClass={cs.LG},
      url={https://arxiv.org/abs/2506.09018}, 
}

@inproceedings{zhang2020cause,
  title={Cause: Learning granger causality from event sequences using attribution methods},
  author={Zhang, Wei and Panum, Thomas and Jha, Somesh and Chalasani, Prasad and Page, David},
  booktitle={International Conference on Machine Learning},
  pages={11235--11245},
  year={2020},
  organization={PMLR}
}

@inproceedings{
psdiff,
title={Unlocking Point Processes through Point Set Diffusion},
author={David L{\"u}dke and Enric Rabasseda Ravent{\'o}s and Marcel Kollovieh and Stephan G{\"u}nnemann},
booktitle={The Thirteenth International Conference on Learning Representations},
year={2025},
url={https://openreview.net/forum?id=4anfpHj0wf}
}

@inproceedings{
addthin,
title={Add and Thin: Diffusion for Temporal Point Processes},
author={David L{\"u}dke and Marin Bilo{\v{s}} and Oleksandr Shchur and Marten Lienen and Stephan G{\"u}nnemann},
booktitle={Thirty-seventh Conference on Neural Information Processing Systems},
year={2023},
url={https://openreview.net/forum?id=tn9Dldam9L}
}

@inproceedings{yuan2023diffstpp,
    author = {Yuan, Yuan and Ding, Jingtao and Shao, Chenyang and Jin, Depeng and Li, Yong},
    title = {Spatio-temporal {D}iffusion {P}oint {P}rocesses},
    year = {2023},
    publisher = {Association for {C}omputing {M}achinery},
    address = {New {Y}ork, {NY}, {USA}},
    booktitle = {Proceedings of the 29th ACM SIGKDD {C}onference on {K}nowledge {D}iscovery and {D}ata {M}ining},
    pages = {3173–3184},
    numpages = {12},
}

@article{li2018learning,
  title={Learning temporal point processes via reinforcement learning},
  author={Li, Shuang and Xiao, Shuai and Zhu, Shixiang and Du, Nan and Xie, Yao and Song, Le},
  journal={Advances in neural information processing systems},
  volume={31},
  year={2018}
}

@article{shchur2021neural,
  title={Neural temporal point processes: A review},
  author={Shchur, Oleksandr and T{\"u}rkmen, Ali Caner and Januschowski, Tim and G{\"u}nnemann, Stephan},
  journal={arXiv preprint arXiv:2104.03528},
  year={2021}
}

@article{xiao2017wasserstein,
  title={Wasserstein learning of deep generative point process models},
  author={Xiao, Shuai and Farajtabar, Mehrdad and Ye, Xiaojing and Yan, Junchi and Song, Le and Zha, Hongyuan},
  journal={Advances in neural information processing systems},
  volume={30},
  year={2017}
}

@inproceedings{zhang2020self,
  title={Self-attentive Hawkes process},
  author={Zhang, Qiang and Lipani, Aldo and Kirnap, Omer and Yilmaz, Emine},
  booktitle={International conference on machine learning},
  pages={11183--11193},
  year={2020},
  organization={PMLR}
}

@inproceedings{shchur2020fast,
  title = {Fast and Flexible Temporal Point Processes with Triangular Maps},
  author = {Shchur, Oleksandr and Gao, Nicholas and Bilo\v{s}, Marin and G{\"u}nnemann, Stephan},
  booktitle={Advances in Neural Information Processing Systems (NeurIPS)},
  year = {2020} 
}

@article{linexploring,
  title={Exploring Generative Neural Temporal Point Process},
  author={Lin, Haitao and Wu, Lirong and Zhao, Guojiang and Pai, Liu and Li, Stan Z},
  journal={Transactions on Machine Learning Research},
  year={2022}
}

@inproceedings{okawa2019deep,
  title={Deep mixture point processes: Spatio-temporal event prediction with rich contextual information},
  author={Okawa, Maya and Iwata, Tomoharu and Kurashima, Takeshi and Tanaka, Yusuke and Toda, Hiroyuki and Ueda, Naonori},
  booktitle={Proceedings of the 25th ACM SIGKDD International Conference on Knowledge Discovery \& Data Mining},
  pages={373--383},
  year={2019}
}

@inproceedings{theis2016note,
  title = {A Note on the Evaluation of Generative Models},
  booktitle = {International {{Conference}} on {{Learning Representations}}},
  author = {Theis, Lucas and van den Oord, A{\"a}ron and Bethge, Matthias},
  year = {2016},
  eprint = {1511.01844},
  publisher = {arXiv},
  doi = {10.48550/arXiv.1511.01844},
  urldate = {2024-11-04}
}

@inproceedings{nichol2021improved,
  title = {Improved {{Denoising Diffusion Probabilistic Models}}},
  booktitle = {International {{Conference}} on {{Machine Learning}}},
  author = {Nichol, Alex and Dhariwal, Prafulla},
  year = {2021},
  eprint = {2102.09672},
  primaryclass = {cs, stat},
  doi = {10.48550/arXiv.2102.09672},
  urldate = {2023-03-12}
}

@article{needleman1970general,
  title = {A General Method Applicable to the Search for Similarities in the Amino Acid Sequence of Two Proteins},
  author = {Needleman, Saul B. and Wunsch, Christian D.},
  year = {1970},
  month = mar,
  journal = {Journal of Molecular Biology},
  volume = {48},
  number = {3},
  pages = {443--453},
  issn = {0022-2836},
  doi = {10.1016/0022-2836(70)90057-4},
  urldate = {2025-09-23}
}

@inproceedings{lienen2024zero,
  title = {From {{Zero}} to {{Turbulence}}: {{Generative Modeling}} for {{3D Flow Simulation}}},
  booktitle = {International {{Conference}} on {{Learning Representations}}},
  author = {Lienen, Marten and L{\"u}dke, David and {Hansen-Palmus}, Jan and G{\"u}nnemann, Stephan},
  year = {2024},
  copyright = {All rights reserved}
}

@inproceedings{heusel2017gans,
  title = {{{GANs Trained}} by a {{Two Time-Scale Update Rule Converge}} to a {{Local Nash Equilibrium}}},
  booktitle = {Neural {{Information Processing Systems}}},
  author = {Heusel, Martin and Ramsauer, Hubert and Unterthiner, Thomas and Nessler, Bernhard and Hochreiter, Sepp},
  year = {2017},
  urldate = {2023-05-13}
}

@misc{touvron2023llama,
  title = {{{LLaMA}}: {{Open}} and {{Efficient Foundation Language Models}}},
  shorttitle = {{{LLaMA}}},
  author = {Touvron, Hugo and Lavril, Thibaut and Izacard, Gautier and Martinet, Xavier and Lachaux, Marie-Anne and Lacroix, Timoth{\'e}e and Rozi{\`e}re, Baptiste and Goyal, Naman and Hambro, Eric and Azhar, Faisal and Rodriguez, Aurelien and Joulin, Armand and Grave, Edouard and Lample, Guillaume},
  year = {2023},
  month = feb,
  number = {arXiv:2302.13971},
  eprint = {2302.13971},
  primaryclass = {cs},
  publisher = {arXiv},
  doi = {10.48550/arXiv.2302.13971},
  urldate = {2025-09-24}
}

@inproceedings{vaswani2017attention,
  title = {Attention {{Is All You Need}}},
  booktitle = {Neural {{Information Processing Systems}}},
  author = {Vaswani, Ashish and Shazeer, Noam and Parmar, Niki and Uszkoreit, Jakob and Jones, Llion and Gomez, Aidan N. and Kaiser, Lukasz and Polosukhin, Illia},
  year = {2017}
}

@misc{dong2024flex,
  title = {Flex {{Attention}}: {{A Programming Model}} for {{Generating Optimized Attention Kernels}}},
  shorttitle = {Flex {{Attention}}},
  author = {Dong, Juechu and Feng, Boyuan and Guessous, Driss and Liang, Yanbo and He, Horace},
  year = {2024},
  month = dec,
  number = {arXiv:2412.05496},
  eprint = {2412.05496},
  primaryclass = {cs},
  publisher = {arXiv},
  doi = {10.48550/arXiv.2412.05496},
  urldate = {2025-09-24}
}

@misc{kingma2023variational,
  title = {Variational {{Diffusion Models}}},
  author = {Kingma, Diederik P. and Salimans, Tim and Poole, Ben and Ho, Jonathan},
  year = {2023},
  month = apr,
  number = {arXiv:2107.00630},
  eprint = {2107.00630},
  primaryclass = {cs, stat},
  publisher = {arXiv},
  doi = {10.48550/arXiv.2107.00630},
  urldate = {2024-08-02}
}

@misc{lienen2025generative,
  title = {Generative {{Modeling}} with {{Bayesian Sample Inference}}},
  author = {Lienen, Marten and Kollovieh, Marcel and G{\"u}nnemann, Stephan},
  year = {2025},
  eprint = {2502.07580},
  primaryclass = {cs.LG}
}
\bibliographystyle{iclr2026_conference}

\appendix

% Group the appendix sections in the PDF bookmarks
\bookmarksetupnext{level=part}
\pdfbookmark{Appendix}{appendix}

% Label appendix sections as Appendix A etc.
\crefalias{section}{appendix}
\crefalias{subsection}{appendix}

\newpage
\section{Model Parameters}\label{sec:model-parameters}

\begin{table}[ht]
	\centering
	\caption{Hyperparameters of our $\model(\cdot \mid \vx_{s})$ model shared across all datasets.}
	\begin{tabular}{@{}ll@{}}
		\toprule
		Parameter                             & Value                                   \\
		\midrule
		Number of $\ins$ bins $\insbins$      & 64                                      \\
		Number of $\sub$ bins $\subbins$      & 64                                      \\
		Maximum $\sub$ distance $\delta$      & $\nicefrac{T}{100}$                     \\
		Maximum log-rate $\lambda_{\text{M}}$ & 32                                      \\
		$\kappa(s)$                           & $1 - {\cos\big(\frac{\pi}{2}s \big)}^2$ \\
		\midrule
		\multicolumn{2}{l}{\textbf{Llama architecture:}}                                \\
		\quad Hidden size $H$                 & 64                                      \\
		\quad Layers                          & 2                                       \\
		\quad Attention heads                 & 4                                       \\
		\midrule
		Optimizer                             & Adam                                    \\
		Sample steps                          & 100                                     \\
		\bottomrule
	\end{tabular}
\end{table}

All \glspl{mlp} have input and output sizes of $H$, except for the final \gls{mlp} whose output size is determined by the number of $\lambda$ and $Q$ parameters of the rate.
The \glspl{mlp} have a single hidden layer of size $4H$.
The sinusoidal embeddings map a scalar $s \in [0, 1]$ to a vector of length $H$.
In contrast to \citet{editflow}, we choose a cosine $\kappa$ schedule $\kappa(s) = 1 - {\cos\big(\frac{\pi}{2}s \big)}^2$ as proposed by \citet{nichol2021improved} for diffusion models as it improved results slightly compared $\kappa(s) = s^3$.

For evaluation, we use an \gls{ema} of the model weights.
We also use low-discrepancy sampling of $s$ in \cref{eq:our-bregman-divergence} during training to smooth the loss and thus training signal \citep{kingma2023variational,lienen2025generative}.

We train all models for \num{20000} steps and select the best checkpoint by its $W_{1}$-over-$\diet$, which we evaluate on a validation set every \num{1000} steps.

\section{Data}

\subsection{Synthetic Datasets}

The six synthetic datasets were generated by \cite{shchur2020fast} following the simulation procedures detailed in Section 4.1 of \cite{omi2019fully}. Each dataset contains 1,000 sequences supported on the interval $T=[0,100]$. They cover a diverse set of temporal dynamics, defined as follows:

\paragraph{Hawkes Processes (H1, H2).} Hawkes processes capture self-exciting features of temporal point processes. The two Hawkes processes are parameterized as follows:
\[
	\lambda(t \mid \mathcal{H}_t)
	=
	\mu
	\;+\;
	\sum_{t_i < t}
	\sum_{j=1}^{M}
	\alpha_j \beta_j
	\exp\{-\beta_j (t - t_i)\},
\]

\noindent
with \textbf{H1} ($M=1$, $\mu=0.2$, $\alpha_1=0.8$, $\beta_1=1.0$) and \textbf{H2} ($M=2$, $\mu=0.2$, $\alpha_1=0.4$, $\beta_1=1.0$, $\alpha_2=0.4$, $\beta_2=20.0$).

\paragraph{Non-stationary Poisson Process (NSP).}
A periodic time-varying intensity:
\[
	\lambda(t \mid \mathcal{H}_t) = 0.99 \sin\!\left(\frac{2\pi t}{20000}\right) + 1.
\]

\paragraph{Stationary Renewal Process (SR).}
Inter-event times $\tau_i = t_{i+1}-t_i$ are i.i.d. from a log-normal distribution (mean $1.0$, std.\ $6.0$):
this produces bursty patterns with short activity bursts followed by long silent periods.

\paragraph{Non-stationary Renewal Process (NSR).}
A stationary renewal process is first generated using a gamma distribution (mean $1.0$, std.\ $0.5$),
then timestamps are time-warped by
\[
	t'_i = \int_{0}^{t_i} r(s)\, ds,
	\qquad
	r(t) = 0.99 \sin\!\left(\frac{2\pi t}{20000}\right) + 1.
\]
This induces temporally varying expected inter-event intervals while preserving local correlations.

\paragraph{Self-correcting Process (SC).}
The intensity grows with the time elapsed since the last event:
\[
	\lambda(t \mid \mathcal{H}_t) = \exp\!\left(t - \sum_{t_i < t} 1\right).
\]
This discourages extended silent periods and promotes regular spacing.

\subsection{Real-World Datasets}

We use the seven real-world datasets proposed by \citep{shchur2020fast}:

\textbf{PG (PUBG)} represents death-event timestamps from matches of PUBG. \textbf{R/C (Reddit-Comments)} consists of comment timestamps within the first 24 hours of threads posted on \texttt{r/askscience},  covering 01.01.2018--31.12.2019. \textbf{R/P (Reddit-Submissions)} captures daily submission timestamps from \texttt{r/politics},  covering 01.01.2017--31.12.2019. \textbf{Tx (Taxi)} are taxi pick-up events in the southern part of Manhattan, New York. \textbf{Tw (Twitter)} covers tweet timestamps of user ID 25073877, collected over multiple years. \textbf{Y/A (Yelp-Airport)} consists of check-in events at McCarran International Airport (27 users, year 2018). Lastly, \textbf{Y/M (Yelp-Mississauga)} presents check-ins for businesses in the city of Mississauga (27 users, year 2018).

\begin{table}[t]
	\centering
	\caption{Summary statistics for all synthetic and real-world datasets. $\boldsymbol{\tau}$ is the average inter-event time.}
	\begin{tabular}{l l c c c c}
		\toprule
		\textbf{Full Name}           & \textbf{Abbrev.} & \textbf{\# Seq.} & \textbf{Mean Len.} & \textbf{Support [0, T]} & $\boldsymbol{\tau}$ \\
		\midrule
		Hawkes 1                     & H1               & 1000             & 95.4               & 100                     & $1.01 \pm 2.38$     \\
		Hawkes 2                     & H2               & 1000             & 97.2               & 100                     & $0.98 \pm 2.56$     \\
		Nonstationary Poisson        & NSP              & 1000             & 100.3              & 100                     & $0.99 \pm 2.22$     \\
		Nonstationary Renewal        & NSR              & 1000             & 98.0               & 100                     & $0.98 \pm 1.83$     \\
		Self-Correcting              & SC               & 1000             & 100.2              & 100                     & $0.99 \pm 0.71$     \\
		Stationary Renewal           & SR               & 1000             & 109.2              & 100                     & $0.83 \pm 2.76$     \\
		\midrule
		PUBG                         & PG               & 3001             & 76.5               & 38 minutes              & $0.41 \pm 0.56$     \\
		Reddit Comments              & R/C              & 1356             & 295.7              & 24 hours                & $0.07 \pm 0.28$     \\
		Reddit Submissions           & R/P              & 1094             & 1129.0             & 24 hours                & $0.02 \pm 0.03$     \\
		Taxi Pick-ups (Manhattan)    & Tx               & 182              & 98.4               & 24 hours                & $0.24 \pm 0.40$     \\
		Twitter Activity             & Tw               & 2019             & 14.9               & 24 hours                & $1.26 \pm 2.80$     \\
		Yelp Check-ins (Airport)     & Y/A              & 319              & 30.5               & 24 hours                & $0.77 \pm 1.10$     \\
		Yelp Check-ins (Mississauga) & Y/M              & 319              & 55.2               & 24 hours                & $0.43 \pm 0.96$     \\
		\bottomrule
	\end{tabular}
	\label{tab:datasets}
\end{table}

\section{Metrics}\label{sec:metrics}

A standard way in generative modeling to compare generated and real data is the Wasserstein distance \citep{heusel2017gans}.
It is the minimum average distance between elements of the two datasets under the optimal (partial) assignment between them,
\begin{equation}
	W_p(\setgenerated, \setdata) = {\Big(\min\nolimits_{\gamma \in \Gamma(\setgenerated, \setdata)} \E\nolimits_{(\vx, \vx') \sim \gamma}\,\big[{d(\vx, \vx')}^p\big]\Big)}^{\nicefrac{1}{p}}
\end{equation}
where $d$ is a distance that compares elements from the two sets.
In the case of sequences of unequal length, one can choose $d$ itself as a nested Wasserstein distance \citep{lienen2024zero}.
\citet{xiao2017wasserstein} were the first to design such a distance between \glspl{tpp}.
They exploit a special case of $W_1$ for sorted sequences of equal length and assign the remaining events of the longer sequence to pseudo-events at $T$ to define
\begin{equation}
	\dxiao(\vx, \vx') = \sum\nolimits_{i = 1}^{|\vx|} |t^{(i)} - t'^{(i)}| + \sum\nolimits_{i=|\vx|+1}^{|\vx'|} |T - t'^{(i)}|
\end{equation}
where $\vx'$ is assumed to be the longer sequence.
$\dxiao$ captures a difference in both location and number of events between two sequences through its two terms.

\citep{shchur2020fast} propose to compute the \gls{mmd} between sets based on a Gaussian kernel and $\dxiao$.
In addition, we evaluate the event count distributions via a Wasserstein-1 distance with respect to a difference in event counts $W_{1,\dlen}$ where $\dlen(\vx, \vx') = \big||\vx| - |\vx'|\big|$.
Finally, we the distributions of inter-event times between our generated sequences and real sequences in $W_{1,\diet}$, i.e., a Wasserstein-1 distance of $\diet$.
$\diet$ is itself the $W_2$ distance between inter-event times of two sequences and quantifies how adjacent events relate to each other to capture more complex patterns.

% On the sequence level, we use $W_2$ distances to prefer close matches, while on the set level, we choose $W_1$ distances to be more robust against outliers on datasets with small test sets.

\section{Ablations}\label{sec:ablations}

We ablate the hyperparameters $\delta$, $\insbins$ and $\subbins$ in \cref{fig:ablation-delta,fig:ablation-ins-bins,fig:ablation-sub-bins,fig:ablation-both-bins}.

\begin{figure*}[!h]
	\centering
	\includegraphics{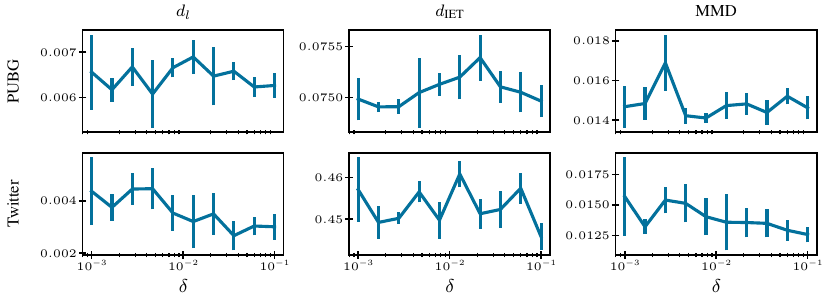}
	\caption{Mean and standard error of $\dlen$, $\diet$ and MMD on two datasets as we vary the $\delta$ parameter for substitutions.}\label{fig:ablation-delta}
\end{figure*}
\begin{figure*}[!h]
	\centering
	\includegraphics{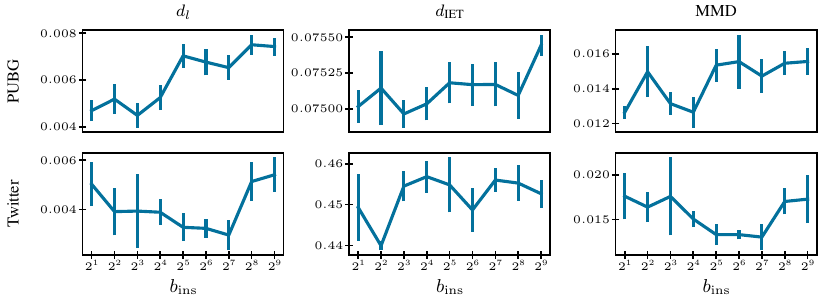}
	\caption{Mean and standard error of $\dlen$, $\diet$ and MMD on two datasets as we vary the number of insertion bins $\insbins$.}\label{fig:ablation-ins-bins}
\end{figure*}
\begin{figure*}[!h]
	\centering
	\includegraphics{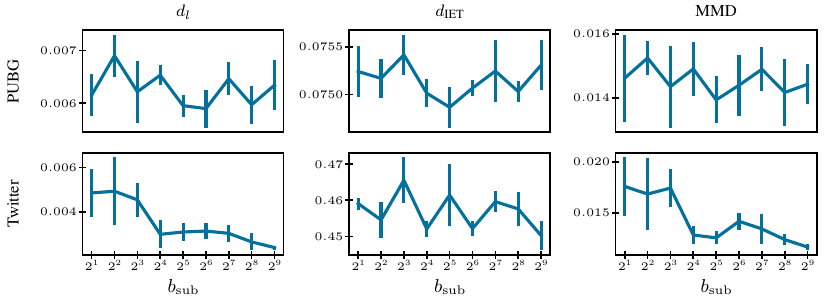}
	\caption{Mean and standard error of $\dlen$, $\diet$ and MMD on two datasets as we vary the number of substitution bins $\subbins$.}\label{fig:ablation-sub-bins}
\end{figure*}
\begin{figure*}[!h]
	\centering
	\includegraphics{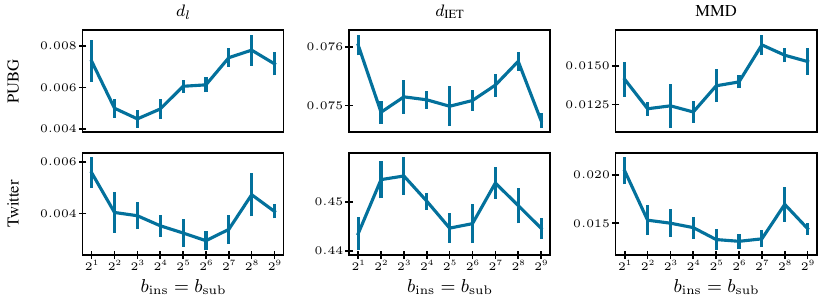}
	\caption{Mean and standard error of $\dlen$, $\diet$ and MMD on two datasets as we vary the number of insertion and substitution bins together.}\label{fig:ablation-both-bins}
\end{figure*}

\newpage

\section{Detailed Results}

\subsection{Inpainting vs.\ Forecasting}

To demonstrate the flexibility of \ours{} for conditional generation, we evaluate its performance when generating events on the interval $[T/3,\, 2T/3]$. In the \emph{forecasting} setting, the model is conditioned only on events occurring before $T/3$, whereas in the \emph{inpainting} setting, it is conditioned on both the past ($t < T/3$) and the future ($t > 2T/3$). As the results show, providing both past and future context substantially improves the quality of the generated middle segment compared to conditioning on the past alone.

\resizebox{\textwidth}{!}{
	\begin{tabular}{lccccccc }
		\toprule
		            & PG          & R/C          & R/P          & Tx          & Tw          & Y/A         & Y/M         \\
		\midrule
		$d_{\text{xiao}}$                                                                                               \\
		Inpainting  & 2.22 ± 0.04 & 22.66 ± 0.77 & 13.13 ± 0.35 & 3.02 ± 0.20 & 1.58 ± 0.03 & 0.59 ± 0.02 & 1.11 ± 0.04 \\
		Forecasting & 2.27 ± 0.05 & 25.53 ± 1.16 & 18.09 ± 0.81 & 3.27 ± 0.33 & 1.65 ± 0.04 & 0.63 ± 0.03 & 1.13 ± 0.05 \\
		\midrule
		\addlinespace
		\text{MRE}                                                                                                      \\
		Inpainting  & 0.29 ± 0.01 & 5.79 ± 0.96  & 0.20 ± 0.01  & 0.60 ± 0.22 & 1.96 ± 0.09 & 0.48 ± 0.01 & 0.74 ± 0.11 \\
		Forecasting & 0.30 ± 0.01 & 5.07 ± 1.13  & 0.33 ± 0.01  & 0.81 ± 0.13 & 2.05 ± 0.11 & 0.52 ± 0.05 & 0.78 ± 0.08 \\
		\midrule
		\addlinespace
		$d_{\text{IET}}$                                                                                                \\
		Inpainting  & 0.39 ± 0.01 & 0.40 ± 0.01  & 0.01 ± 0.00  & 0.10 ± 0.01 & 1.13 ± 0.02 & 0.80 ± 0.05 & 0.70 ± 0.06 \\
		Forecasting & 0.41 ± 0.01 & 0.44 ± 0.03  & 0.02 ± 0.00  & 0.10 ± 0.00 & 1.18 ± 0.01 & 0.78 ± 0.07 & 0.72 ± 0.08 \\
		\addlinespace
		\bottomrule
	\end{tabular}}

\subsection{Parametric TPP samples}

To illustrate how well each model captures parametric TPPs, we draw 200 samples for the Hawkes and Self-Correcting processes. In \cref{fig:samples_count}, we plot the cumulative count $N(t)$ for each sample, while \cref{fig:samples_event} shows each event sequence as a separate row, directly visualizing the events over time. These visualizations further highlight the strong unconditional sampling performance of \ours{} demonstrated in \cref{tab:conditional}.

\begin{figure*}[t]
	\centering
	\includegraphics{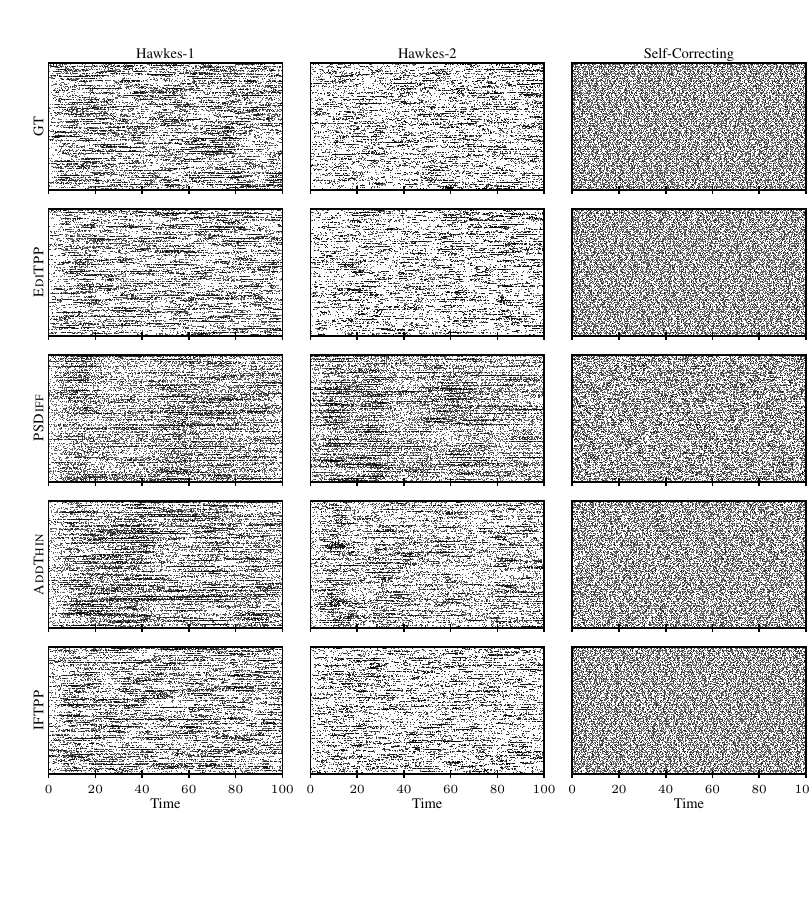}
	\caption{Event times for 200 samples from ground truth data (GT) and each model. Each event sequence is represented as a separate row.}\label{fig:samples_event}
\end{figure*}

\begin{figure*}[t]
	\centering
	\includegraphics{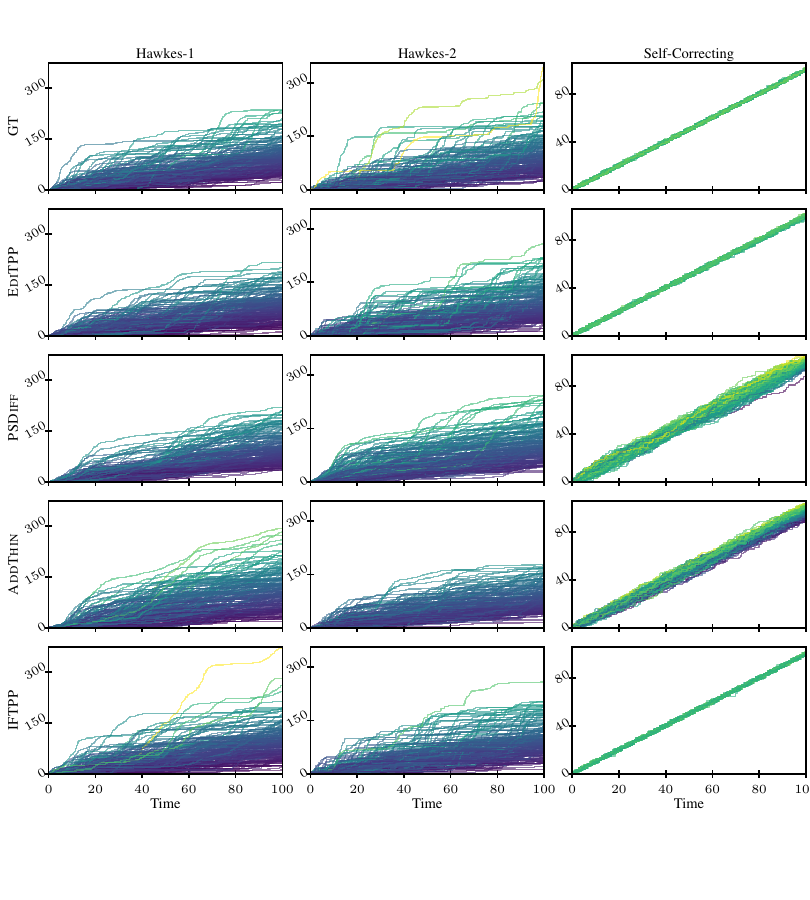}
	\caption{$N(t)$ for 200 samples from ground truth data (GT) and each model.}\label{fig:samples_count}
\end{figure*}

\newpage
\subsection{Full results}\label{sec:tables}
\begin{table}[ht]
\centering
\caption{Forecasting accuracy up to $T$ measured by $\diet$.}
\begin{tabular}{lllll}
\toprule
{} & {\ours{}} & {\psdiff} & {\addthin} & {\iftpp} \\
\midrule
PUBG & \textbf{\num{0.400+-0.002}} & \underline{\num{0.413+-0.009}} & \underline{\num{0.403+-0.010}} & \num{0.473+-0.019} \\
Reddit Comments & \underline{\num{0.684+-0.005}} & \textbf{\num{0.625+-0.012}} & \underline{\num{0.693+-0.012}} & \underline{\num{0.684+-0.012}} \\
Reddit Posts & \underline{\num{0.010+-0.000}} & \textbf{\num{0.009+-0.000}} & \underline{\num{0.010+-0.001}} & \num{0.015+-0.003} \\
Taxi & \textbf{\num{0.113+-0.003}} & \textbf{\num{0.113+-0.001}} & \underline{\num{0.116+-0.001}} & \num{0.145+-0.009} \\
Twitter & \textbf{\num{1.441+-0.020}} & \underline{\num{1.487+-0.012}} & \underline{\num{1.493+-0.033}} & \num{2.187+-0.029} \\
Yelp Airport & \underline{\num{0.497+-0.009}} & \textbf{\num{0.492+-0.005}} & \textbf{\num{0.493+-0.013}} & \num{0.587+-0.019} \\
Yelp Mississauga & \underline{\num{0.272+-0.003}} & \textbf{\num{0.262+-0.003}} & \textbf{\num{0.260+-0.003}} & \num{0.388+-0.024} \\
\bottomrule
\end{tabular}

\end{table}

\begin{table}[ht]
\centering
\caption{Forecasting accuracy up to $T$ measured by mean relative error of event counts.}
\begin{tabular}{lllll}
\toprule
{} & {\ours{}} & {\psdiff} & {\addthin} & {\iftpp} \\
\midrule
PUBG & \underline{\num{0.349+-0.001}} & \textbf{\num{0.339+-0.008}} & \num{0.367+-0.005} & \num{3.892+-0.035} \\
Reddit Comments & \underline{\num{3.594+-0.118}} & \textbf{\num{3.260+-0.268}} & \num{14.777+-3.226} & \num{7.515+-2.112} \\
Reddit Posts & \textbf{\num{0.281+-0.001}} & \underline{\num{0.296+-0.006}} & \num{0.457+-0.065} & \num{0.352+-0.022} \\
Taxi & \num{1.234+-0.036} & \num{1.140+-0.043} & \textbf{\num{0.301+-0.014}} & \underline{\num{0.321+-0.018}} \\
Twitter & \underline{\num{2.327+-0.042}} & \num{2.435+-0.106} & \num{2.984+-0.246} & \textbf{\num{2.060+-0.027}} \\
Yelp Airport & \underline{\num{0.350+-0.007}} & \textbf{\num{0.346+-0.004}} & \textbf{\num{0.347+-0.014}} & \num{0.366+-0.009} \\
Yelp Mississauga & \num{0.902+-0.027} & \num{0.920+-0.033} & \textbf{\num{0.374+-0.012}} & \underline{\num{0.392+-0.012}} \\
\bottomrule
\end{tabular}

\end{table}

\begin{table}[ht]
\centering
\caption{Forecasting accuracy up to $T$ measured by $\dxiao$.}
\begin{tabular}{lllll}
\toprule
{} & {\ours{}} & {\psdiff} & {\addthin} & {\iftpp} \\
\midrule
PUBG & \underline{\num{2.478+-0.007}} & \textbf{\num{2.400+-0.007}} & \underline{\num{2.466+-0.024}} & \num{5.954+-0.195} \\
Reddit Comments & \underline{\num{34.135+-0.382}} & \textbf{\num{32.467+-0.534}} & \num{87.666+-20.184} & \num{39.010+-7.508} \\
Reddit Posts & \textbf{\num{48.776+-0.355}} & \textbf{\num{47.829+-1.050}} & \underline{\num{72.754+-12.134}} & \underline{\num{63.256+-9.695}} \\
Taxi & \underline{\num{4.464+-0.088}} & \underline{\num{4.444+-0.076}} & \textbf{\num{4.032+-0.129}} & \num{4.744+-0.125} \\
Twitter & \underline{\num{2.669+-0.022}} & \underline{\num{2.635+-0.078}} & \num{2.802+-0.132} & \textbf{\num{2.557+-0.055}} \\
Yelp Airport & \textbf{\num{1.524+-0.013}} & \textbf{\num{1.512+-0.016}} & \underline{\num{1.548+-0.026}} & \num{1.795+-0.015} \\
Yelp Mississauga & \underline{\num{3.027+-0.046}} & \underline{\num{3.005+-0.046}} & \textbf{\num{2.895+-0.039}} & \num{3.430+-0.047} \\
\bottomrule
\end{tabular}

\end{table}

\begin{table}[ht]
\centering
\caption{Sample quality as measured by $\operatorname{MMD}$.}
\begin{tabular}{lllll}
\toprule
{} & {\ours{}} & {\psdiff} & {\addthin} & {\iftpp} \\
\midrule
Hawkes-1 & \textbf{\num{0.011+-0.002}} & \num{0.033+-0.009} & \num{0.024+-0.009} & \underline{\num{0.016+-0.002}} \\
Hawkes-2 & \textbf{\num{0.012+-0.001}} & \underline{\num{0.018+-0.006}} & \underline{\num{0.018+-0.006}} & \textbf{\num{0.012+-0.001}} \\
Nonstationary Poisson & \textbf{\num{0.017+-0.003}} & \textbf{\num{0.020+-0.005}} & \underline{\num{0.035+-0.011}} & \underline{\num{0.032+-0.008}} \\
Nonstationary Renewal & \textbf{\num{0.035+-0.001}} & \num{0.059+-0.006} & \num{0.157+-0.084} & \underline{\num{0.039+-0.007}} \\
PUBG & \textbf{\num{0.014+-0.001}} & \underline{\num{0.032+-0.012}} & \num{0.046+-0.025} & \num{0.162+-0.010} \\
Reddit Comments & \textbf{\num{0.008+-0.001}} & \textbf{\num{0.006+-0.002}} & \underline{\num{0.063+-0.012}} & \textbf{\num{0.007+-0.003}} \\
Reddit Posts & \underline{\num{0.024+-0.001}} & \textbf{\num{0.010+-0.002}} & \num{0.102+-0.004} & \underline{\num{0.020+-0.007}} \\
Self-Correcting & \textbf{\num{0.077+-0.004}} & \underline{\num{0.198+-0.002}} & \num{0.246+-0.018} & \textbf{\num{0.067+-0.011}} \\
Stationary Renewal & \textbf{\num{0.010+-0.002}} & \underline{\num{0.024+-0.005}} & \underline{\num{0.025+-0.013}} & \textbf{\num{0.012+-0.002}} \\
Taxi & \textbf{\num{0.031+-0.002}} & \underline{\num{0.038+-0.005}} & \underline{\num{0.041+-0.004}} & \num{0.050+-0.003} \\
Twitter & \textbf{\num{0.013+-0.002}} & \num{0.034+-0.007} & \num{0.044+-0.012} & \underline{\num{0.026+-0.005}} \\
Yelp Airport & \textbf{\num{0.037+-0.002}} & \underline{\num{0.041+-0.004}} & \num{0.118+-0.036} & \num{0.058+-0.002} \\
Yelp Mississauga & \underline{\num{0.040+-0.003}} & \underline{\num{0.034+-0.007}} & \underline{\num{0.037+-0.006}} & \textbf{\num{0.029+-0.002}} \\
\bottomrule
\end{tabular}

\end{table}

\begin{table}[ht]
\centering
\caption{Sample quality as measured by $W_1$-over-$\diet$.}
\begin{tabular}{lllll}
\toprule
{} & {\ours{}} & {\psdiff} & {\addthin} & {\iftpp} \\
\midrule
Hawkes-1 & \textbf{\num{0.526+-0.020}} & \num{0.865+-0.035} & \underline{\num{0.655+-0.081}} & \underline{\num{0.628+-0.030}} \\
Hawkes-2 & \textbf{\num{0.546+-0.005}} & \num{0.991+-0.038} & \num{0.703+-0.049} & \underline{\num{0.582+-0.009}} \\
Nonstationary Poisson & \textbf{\num{0.306+-0.005}} & \textbf{\num{0.303+-0.007}} & \underline{\num{0.318+-0.015}} & \underline{\num{0.317+-0.006}} \\
Nonstationary Renewal & \textbf{\num{0.224+-0.006}} & \num{0.511+-0.016} & \underline{\num{0.393+-0.064}} & \textbf{\num{0.229+-0.027}} \\
PUBG & \textbf{\num{0.075+-0.000}} & \num{0.090+-0.001} & \underline{\num{0.080+-0.003}} & \num{0.303+-0.039} \\
Reddit Comments & \textbf{\num{0.144+-0.003}} & \underline{\num{0.157+-0.006}} & \num{0.532+-0.014} & \num{0.176+-0.008} \\
Reddit Posts & \underline{\num{0.006+-0.000}} & \textbf{\num{0.004+-0.000}} & \num{0.020+-0.001} & \num{0.007+-0.001} \\
Self-Correcting & \textbf{\num{0.064+-0.000}} & \num{0.326+-0.003} & \num{0.151+-0.005} & \underline{\num{0.065+-0.001}} \\
Stationary Renewal & \textbf{\num{0.697+-0.018}} & \num{1.281+-0.049} & \underline{\num{0.941+-0.145}} & \textbf{\num{0.714+-0.028}} \\
Taxi & \underline{\num{0.111+-0.001}} & \underline{\num{0.111+-0.001}} & \textbf{\num{0.088+-0.003}} & \num{0.174+-0.015} \\
Twitter & \textbf{\num{0.460+-0.004}} & \num{0.672+-0.007} & \num{0.545+-0.024} & \underline{\num{0.492+-0.023}} \\
Yelp Airport & \textbf{\num{0.246+-0.002}} & \textbf{\num{0.244+-0.004}} & \underline{\num{0.316+-0.046}} & \underline{\num{0.318+-0.017}} \\
Yelp Mississauga & \textbf{\num{0.226+-0.003}} & \textbf{\num{0.225+-0.003}} & \underline{\num{0.236+-0.004}} & \num{0.276+-0.017} \\
\bottomrule
\end{tabular}

\end{table}

\begin{table}[ht]
\centering
\caption{Sample quality as measured by $W_1$-over-$\dlen$.}
\begin{tabular}{lllll}
\toprule
{} & {\ours{}} & {\psdiff} & {\addthin} & {\iftpp} \\
\midrule
Hawkes-1 & \textbf{\num{0.008+-0.001}} & \num{0.027+-0.008} & \num{0.033+-0.015} & \underline{\num{0.020+-0.004}} \\
Hawkes-2 & \textbf{\num{0.007+-0.001}} & \num{0.030+-0.009} & \num{0.022+-0.014} & \underline{\num{0.013+-0.003}} \\
Nonstationary Poisson & \textbf{\num{0.003+-0.001}} & \underline{\num{0.006+-0.001}} & \num{0.013+-0.005} & \num{0.012+-0.003} \\
Nonstationary Renewal & \textbf{\num{0.001+-0.000}} & \underline{\num{0.013+-0.001}} & \num{0.049+-0.022} & \num{0.014+-0.011} \\
PUBG & \textbf{\num{0.006+-0.000}} & \underline{\num{0.016+-0.008}} & \underline{\num{0.024+-0.014}} & \num{0.295+-0.007} \\
Reddit Comments & \underline{\num{0.019+-0.002}} & \textbf{\num{0.013+-0.003}} & \num{0.370+-0.081} & \num{0.039+-0.023} \\
Reddit Posts & \num{0.057+-0.003} & \textbf{\num{0.025+-0.003}} & \num{0.336+-0.045} & \underline{\num{0.032+-0.011}} \\
Self-Correcting & \textbf{\num{0.001+-0.000}} & \underline{\num{0.011+-0.001}} & \num{0.023+-0.002} & \textbf{\num{0.001+-0.001}} \\
Stationary Renewal & \textbf{\num{0.006+-0.002}} & \num{0.030+-0.019} & \num{0.042+-0.022} & \underline{\num{0.023+-0.005}} \\
Taxi & \textbf{\num{0.025+-0.002}} & \textbf{\num{0.028+-0.004}} & \textbf{\num{0.023+-0.006}} & \underline{\num{0.029+-0.003}} \\
Twitter & \textbf{\num{0.003+-0.001}} & \underline{\num{0.006+-0.003}} & \num{0.015+-0.008} & \underline{\num{0.007+-0.002}} \\
Yelp Airport & \textbf{\num{0.014+-0.002}} & \textbf{\num{0.015+-0.004}} & \num{0.060+-0.021} & \underline{\num{0.033+-0.003}} \\
Yelp Mississauga & \underline{\num{0.017+-0.003}} & \textbf{\num{0.015+-0.002}} & \textbf{\num{0.016+-0.003}} & \num{0.025+-0.006} \\
\bottomrule
\end{tabular}

\end{table}
\begin{table}[!h]
	\centering
	\caption{Average number of edit operations during unconditional sampling.}
	\label{tab:edit_ops_full}
	\begin{tabular}{l|ccc|cc}
		\toprule
		      & \multicolumn{3}{c|}{\ours}  & \multicolumn{2}{c}{\psdiff}                                                  \\
		\cmidrule(lr){2-4} \cmidrule(lr){5-6}
		      & Ins                         & Del                         & Sub          & Ins             & Del           \\
		\midrule
		H1    & 55.05 ± 28.6                & 65.93 ± 10.5                & 37.74 ± 10.6 & 92.58 ± 34.3    & 102.50 ± 10.6 \\
		H2    & 63.69 ± 32.4                & 71.72 ± 9.6                 & 30.98 ± 8.7  & 97.57 ± 38.9    & 101.30 ± 10.2 \\
		NSP   & 49.59 ± 7.3                 & 49.66 ± 7.1                 & 50.56 ± 8.2  & 100.14 ± 9.7    & 100.16 ± 9.8  \\
		NSR   & 42.39 ± 5.8                 & 44.23 ± 7.2                 & 55.88 ± 7.8  & 97.58 ± 7.5     & 100.51 ± 10.5 \\
		PG    & 56.69 ± 7.2                 & 19.71 ± 4.4                 & 19.96 ± 4.8  & 76.32 ± 8.7     & 40.90 ± 6.4   \\
		R/C   & 247.83 ± 251.3              & 8.69 ± 6.6                  & 15.38 ± 7.4  & 274.49 ± 254.3  & 24.45 ± 5.7   \\
		R/P   & 972.89 ± 300.2              & 0.11 ± 0.3                  & 23.74 ± 5.0  & 1109.78 ± 307.1 & 24.56 ± 7.4   \\
		SC    & 34.93 ± 5.7                 & 34.37 ± 6.9                 & 66.18 ± 8.6  & 98.95 ± 7.8     & 99.61 ± 10.2  \\
		SR    & 70.67 ± 21.6                & 64.09 ± 11.3                & 38.00 ± 11.5 & 108.85 ± 31.4   & 101.92 ± 10.9 \\
		Tw    & 11.83 ± 9.4                 & 22.34 ± 4.6                 & 3.04 ± 2.3   & 14.37 ± 10.3    & 24.46 ± 5.2   \\
		Tx    & 96.62 ± 14.4                & 9.35 ± 3.3                  & 17.23 ± 4.8  & 97.69 ± 17.4    & 24.37 ± 5.2   \\
		Y/A   & 29.43 ± 6.4                 & 22.64 ± 5.0                 & 9.00 ± 3.4   & 30.42 ± 6.3     & 24.30 ± 4.8   \\
		Y/M   & 54.81 ± 13.8                & 17.14 ± 4.2                 & 11.37 ± 3.8  & 56.54 ± 15.5    & 24.45 ± 4.9   \\
		\midrule
		Mean  & 137.42                      & 33.08                       & 29.16        & 173.48          & 61.04         \\
		\midrule
		Total & \multicolumn{3}{c|}{199.65} & \multicolumn{2}{c}{234.52}                                                   \\
		\bottomrule
	\end{tabular}
\end{table}

\end{document}